\documentclass[10pt,twocolumn,letterpaper]{article}

\usepackage{cvpr}
\usepackage{times}
\usepackage{epsfig}
\usepackage{graphicx}
\usepackage{amsmath}
\usepackage{amssymb}
\usepackage{subcaption}
\usepackage{mathtools}
\usepackage{footmisc}
\DeclareMathOperator{\sign}{sgn}
\usepackage[pagebackref=true,breaklinks=true,letterpaper=true,colorlinks,bookmarks=false]{hyperref}

 \cvprfinalcopy 
\def\cvprPaperID{XXXX} 

\ifcvprfinal\fi

\usepackage[utf8]{inputenc}
\usepackage[T1]{fontenc}
\usepackage[dvipsnames]{xcolor}
\usepackage{algorithm}
\usepackage[noend]{algpseudocode}
\usepackage{graphicx}
\usepackage{booktabs}
\usepackage{multirow}
\usepackage{multicol}
\usepackage{float}

\newif\ifdraft

\newcommand{\ms}[1]{\ifdraft {\color{green}{#1}} \else {#1}\fi}
\newcommand{\kn}[1]{\ifdraft {\color{blue}{#1}} \else {#1}\fi}

\newcommand{\MS}[1]{\ifdraft {\color{green}{\textbf{MS: #1}}}\else {}\fi}
\newcommand{\KN}[1]{\ifdraft {\color{Fuchsia}{\textbf{KN: #1}}}\else {}\fi}

\newcommand{\comment}[1]{}

\newcommand{\Introfigadaptive}[1]{./figures/introfig2/#1}
\newcommand{\expfigadaptivecity}[1]{./figures/adaptivecity/#1}
\newcommand{\expfigunivcity}[1]{./figures/univcity/#1}
\newcommand{\expfigboundarycity}[1]{./figures/boundarycity/#1}
\newcommand{\expfigadaptivevoc}[1]{./figures/adaptivevoc/#1}

\newcommand{\supunivcity}[1]{./figures/Sup_Univ/#1}
\newcommand{\supadaptivecity}[1]{./figures/Sup_Adaptive/#1}
\newcommand{\supboundarycity}[1]{./figures/Sup_boundary/#1}
\newcommand{\supindirectcity}[1]{./figures/Sup_Indirect/#1}
\newcommand{\supvizinternalcity}[1]{./figures/Sup_VizInternal/#1}
\newcommand{\supadaptivevoc}[1]{./figures/Sup_Pascal/#1}
\newcommand{\supdetectcity}[1]{./figures/Sup_detect/#1}

\begin{document}

\title{Indirect Local Attacks for Context-aware Semantic Segmentation Networks}

\author{Krishna Kanth Nakka and Mathieu Salzmann\\
	Computer  Vision Lab, EPFL\\
}

\maketitle
\newcommand{\bp}{\mathbf{\emph{p}}}
\newcommand{\bu}{\mathbf{u}}
\newcommand{\bb}{\mathbf{b}}
\newcommand{\bfs}{\mathbf{\emph{f}}}
\newcommand{\bF}{\mathbf{F}}
\newcommand{\bXhat}{\mathbf{\hat{X}}}
\newcommand{\bM}{\mathbf{M}}
\newcommand{\bX}{\mathbf{X}}
\newcommand{\ba}{\mathbf{a}}
\newcommand{\bc}{\mathbf{c}}
\newcommand{\bt}{\mathbf{t}}
\newcommand{\mASRT}{\text{mASR}_{(\ba,\bt)}}
\newcommand{\mASRU}{\text{mASR}_{(\ba,\bc)}}
\newcommand{\mIoUT}{\text{mIoU}_{(\ba,\bt)}}
\newcommand{\mIoUU}{\text{mIoU}_{(\ba,\bc)}}
\newcommand{\ours}[0]{{\bf Ours}}
\newcommand{\dyn}[0]{{Dyn}}
\newcommand{\up}[0]{{UP }}   
\newcommand{\ap}[0]{{AP }}  
\newcommand{\full}[0]{{Full }}  
\newcommand{\fs}[0]{{FS }}


\begin{abstract}
Recently, deep networks have achieved impressive semantic segmentation performance, in particular thanks to their use of larger contextual information. In this paper, we show that the resulting networks are sensitive not only to global attacks, where perturbations affect the entire input image, but also to indirect local attacks where perturbations are confined to a small image region that does not overlap with the area that we aim to fool. To this end, we introduce several indirect attack strategies, including adaptive local attacks, aiming to find the best image location to perturb, and universal local attacks. Furthermore, we propose attack detection techniques both for the global image level and to obtain a pixel-wise localization of the fooled regions. Our results are unsettling: Because they exploit \kn{a larger} context, more accurate semantic segmentation networks are more sensitive to indirect local attacks.

\comment{
 Recently, adversarial examples which are  well studied in image classification tasks, have been extended to other learning problems.  In this paper, we conduct local adversarial attacks to  modern segmentation networks and show that such attacks can compromise their performance. We show that by corrupting a small patch of less than 1\% of the image size can significantly affect segmentation estimates even beyond the region of the attack. We observed  networks which use using an context information are very sensitive to these attacks, while the non context counterparts  are relatively less affected. We study the impact of various context strategies such as global pooling, dilated convolutions, long-range spatial connections and understand the propagation of perturbation in each  network. Further, we enforce group sparsity prior on patch perturbation to adaptively find the most sensitive regions in the context static classes to fool the dynamic object  pixels.

Also, we present what to our knowledge is the first rigorous evaluation of adversarial attacks detection methods, which are  studied in classification regime, on modern semantic segmentation models both at image and pixel level.
}

\end{abstract}


\section{Introduction}\label{sec:intro}

Deep Neural Networks (DNNs) are highly expressive models and achieve state-of-the-art performance on many computer vision tasks. In particular, the powerful backbones originally developed for image recognition have now be recycled for semantic segmentation, via the development of fully convolutional networks (FCNs)~\cite{fcn2015}. The success of these initial FCNs, however, was impeded by their limited understanding of surrounding context. As such, recent techniques have focused on incorporating contextual information via dilated convolutions~\cite{drn2017}, pooling operations~\cite{parsenet2015,psp2017}, or attention mechanisms~\cite{psanet2018,danet2019}.

Despite this success, recent studies have \comment{evidenced} \kn{shown} that DNNs are vulnerable to adversarial attacks. That is, small, dedicated perturbations to the input images can make a network produce virtually arbitrarily incorrect predictions. While this has been mostly studied in the context of image recognition~\cite{nguyen2015deep, kurakin2016adversarial,dong2018boosting,moosavi2016deepfool,saliencyattack}, a few recent works have nonetheless discussed such adversarial attacks for semantic segmentation~\cite{DAG2017,arnab2018robustness,univperturb}. These methods, however, remain limited to global perturbations to the entire image. Here, we argue that local attacks are more realistic, in that, in practice, they would allow one to modify the physical environment to fool a network. This, in some sense, was the task addressed in~\cite{robustphysical2018}, where stickers were placed on traffic poles so that an image recognition network would misclassify the corresponding traffic signs. In this scenario, however, the attack was directly performed on the targeted object. 


\providecommand{\localwidth}{}
\renewcommand{\localwidth}{0.45\linewidth}

\providecommand{\localheight}{}
\renewcommand{\localheight}{1.5cm}

\begin{figure}[t]
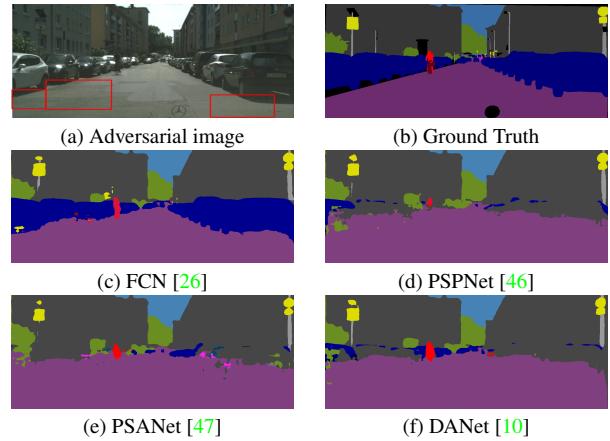

	\footnotesize
	\centering
	\begin{tabular}{ccc}

		\includegraphics[height= \localheight,width=\localwidth]{\Introfigadaptive{94_image1.png}}&
		\includegraphics[height= \localheight,width=\localwidth]{\Introfigadaptive{94_GT.png}}\\
		(a) Adversarial image &  (b) Ground Truth \\

		\includegraphics[height= \localheight,width=\localwidth]{\Introfigadaptive{94_fcn.png}}&
		\includegraphics[height= \localheight,width=\localwidth]{\Introfigadaptive{94_psp.png}}\\
				(c) FCN~\cite{fcn2015} &(d) PSPNet~\cite{psp2017} \\

	\includegraphics[height= \localheight,width=\localwidth]{\Introfigadaptive{94_psa.png}}&
	\includegraphics[height= \localheight,width=\localwidth]{\Introfigadaptive{94_danet.png}}\\
		(e) PSANet~\cite{psanet2018} & (f) DANet~\cite{danet2019} \\	
	\end{tabular}
	\vspace{-3mm}
	\caption{{\bf Indirect Local Attacks.} An adversarial  input image   {\bf(a)} is attacked with an imperceptible noise in local regions, shown as red boxes, to fool the dynamic objects.  Such \emph{indirect} local  attacks barely affects FCN~\cite{fcn2015} {\bf(c)}. By contrast,  modern networks that leverage context to achieve higher accuracy, such as PSPNet~\cite{psp2017} {\bf(d)}, PSANet~\cite{psanet2018} {\bf(e)}  and DANet~\cite{danet2019} {\bf(f)} are more strongly affected, even in regions far away from the perturbed area. 	\KN{(a) is the boxes the from the adaptive indirect attack on PSANet. While for rest of the networks perturbed area almost overlaps but not exactly as on (a)'s. boxes. 
			Will run an experiment fixing (a)'s boxes to other networks. Is that sound okay? or can we show each network's most optimized boxes..}\MS{I think for the teaser it is ok to show just one set of boxes. However, can you remove the grid structure? It looks weird.}
}\label{fig:teaser_adaptive_city}
\end{figure}

Here, by contrast, we study the impact of \emph{indirect} local attacks, where the perturbations are performed on regions outside the targeted objects.\comment{This, for instance, would allow one to place a sticker on the back of their car such that the car behind mislabels the dynamic objects, such as pedestrians, as the nearest background class.}\kn{ This, for instance, would allow one to place a sticker on the building such that the nearby dynamic objects, such as cars and pedestrians, gets mislabeled as the nearest background class.}\KN{ Few people commented that in all our experiments, perturbations happens on static area, sticker on car example is confusing. We can write as, sticker on the road and building such that all dynamic objects mis-classifies.}
To this end, we first investigate the general idea of \emph{indirect} attacks, where the perturbations can occur anywhere in the image except on the targeted objects. We then switch to the more realistic case of \emph{localized} indirect attacks, and design a group sparsity-based strategy to confine the perturbed region to a small area outside of the targeted objects.\KN{From here, universal attack performs an untargeted one. Both local perturbations are intended to do different things. We need to change the sentence 'instead of computing such local perturbations' in the below commented text}\comment{Finally, instead of computing such local perturbations for every individual image, we show that a single universal fixed-size patch can be learned from all training images to attack a given network.}
\kn{In addition, we show the existence of a single universal fixed-size patch that can be learned from all training images to attack an entire unseen image in an untargeted way.}

The conclusions of our experiments are disturbing: In short, more accurate semantic segmentation networks are more sensitive to indirect local attacks.\comment{This is illustrated by Figure~\ref{fig:teaser_adaptive_city} where perturbing a patch covering 0.53\% of the image has a much larger impact on the context-aware PSPNet~\cite{psp2017}, PSANet~\cite{psanet2018} and DANet~\cite{danet2019} than on a simple FCN~\cite{fcn2015}}
This is illustrated by Figure~\ref{fig:teaser_adaptive_city}, where perturbing 
\comment{a patch} \kn{few patches}
in a static region has much larger impact on the dynamic objects for the context-aware PSPNet~\cite{psp2017}, PSANet~\cite{psanet2018} and DANet~\cite{danet2019} than for a simple FCN~\cite{fcn2015}. This, however, has to be expected, because the use of context, which improves segmentation accuracy, also increases the network's receptive field, thus allowing the perturbation to be propagated to more distant image regions.
\label{key}
Motivated by this unsettling sensitivity of segmentation networks to indirect local attacks, we then turn our focus to adversarial attack detection. In contrast to the only two existing works that have tackled attack detection for semantic segmentation~\cite{SC2018,lis2019}, we perform detection not only at the global image level, but locally at the pixel level. Specifically, we introduce an approach to localizing the regions whose predictions were affected by the attack, i.e., not the image regions that were perturbed. In an autonomous driving scenario, this would allow one to focus more directly on the potential dangers themselves, rather than on the image regions that caused them.

To summarize, our contributions are as follows. We introduce the idea of indirect local adversarial attacks for semantic segmentation networks. We design an adaptive, image-dependent local attack strategy. We show the existence of a universal, \kn{non} image-independent \KN{ isn't it non image-dependent? Universal attack is non image dependent} adversarial patch for a given network and dataset. We study the impact of context on a network's sensitivity to our indirect local attacks. We introduce a method to detect indirect local attacks at both image level and pixel level. Our attack and detection code will be made publicly available upon acceptance of this paper.


\section{Related Work}\label{sec:related}

\noindent{ \textbf{Context in Semantic Segmentation Networks.}} While context has been shown to improve the results of traditional semantic segmentation methods~\cite{he2004multiscale,kohli2009robust,krahenbuhl2011efficient,gonfaus2010harmony}, the early deep fully-convolutonal semantic segmentation networks~\cite{fcn2015,hariharan2015hypercolumns} \MS{Are there other papers that we could cite here?}\KN{Added a new one} only gave each pixel a limited receptive field, thus encoding relatively local relationships. Since then, several solutions have been proposed to account for wider context. \comment{ For example, UNet~\cite{unet2015}  increases the receptive field 
by subsequent pooling of the convolutional feature maps, followed \ms{by further up-sampling the intermediate, low-resolution representation back to the input resolution.} \MS{Correct?}\KN{UNet covers larger context than FCNs due to its contracting layers (encoding layers).  UNet's decoder is primarily used for a finer  segmentation map. Receptive field is increased due to its contracting encoding layers than compared to FCNs encoding layers. We can rewrite as, } }
For example, 	UNet~\cite{unet2015}  uses contracting path to capture larger context followed by a expanding path to upsample the intermediate  low-resolution representation back to the input resolution.  ParseNet~\cite{parsenet2015} relies on global pooling of the final convolutional features to aggregate context information. This idea was extended to using different pooling strides in PSPNet~\cite{psp2017}, so as to encode different levels of context. In~\cite{drn2017}, dilated convolutions were introduced to increase the size fo the receptive field. PSANet~\cite{psanet2018} is designed so that each local feature vector is connected to all the other ones in the feature map, thus learning contextual information adaptively. EncNet~\cite{encnet} captures context via a separate network branch that predicts the presence of the object categories in the scene without localizing them. DANet~\cite{danet2019} uses a dual attention mechanism to attend to the most important spatial and channel locations in the final feature map.  In particular, the DANet position attention module selectively aggregates the features at all positions using a weighted sum. In practice, all of these strategies to use larger contextual information have been shown to outperform simple FCNs on clean samples. Here, however, we show that this makes the resulting networks more vulnerable to   indirect local adversarial attacks, even when the perturbed region covers less than 1\% of the input image.

\vspace*{0.1cm}

\noindent{\textbf{Adversarial Attacks on Semantic Segmentation:}} 
Adversarial attacks aim to perturb an input image with an imperceptible noise so as to make a DNN produce erroneous predictions. So far, the main focus of the adversarial attack literature has been image classification, for which diverse attack and defense strategies have been proposed~\cite{goodfellow2014explaining,carlini2017towards,nguyen2015deep,kurakin2016adversarial,dong2018boosting,moosavi2016deepfool,saliencyattack}.  In this context, it was shown that deep networks can be attacked even when one does not have access to the model weights~\cite{liu2016delving,papernot2016transferability}, that attacks can be transferred across different networks~\cite{tramer2017ensemble}, and that universal perturbations that can be applied to any input image exist~\cite{moosavi2017universal,moosavi2017analysis,poursaeed2018generative}.

Motivated by the observations made in the context of image classification, adversarial attacks were extended to semantic segmentation. In~\cite{arnab2018robustness}, the effectiveness of attack strategies designed for classification was studied for different segmentation networks. In~\cite{DAG2017}, a dense adversary generation attack was proposed, consisting of projecting the gradient in each iteration with minimal distortion. In~\cite{univperturb}, a universal perturbation was learnt using the whole image dataset. 
None of these works, however, impose any constraints on the location of the attack in the input image. As such, the entire image is perturbed, which, while effective when the attacker has access to the image itself, would not allow one to physically modify the scene so as to fool, e.g., autonomous vehicles.

This, in essence, was the task addressed in~\cite{robustphysical2018}, where it was shown that placing a small, well-engineered patch on a traffic sign was able to fool a classification network into making wrong decisions. Such attacks, however, are \emph{direct}, in the sense that the perturbation is located on the object that should be misclassified. Here, by contrast, we study the impact of \emph{indirect} local attacks, where the perturbation is outside the object of interest. This would allow one to modify static portions of the scene so as to, e.g., make dynamic objects disappear. We then study the impact of the contextual information exploited by different network architectures on robustness, and introduce an attack strategy that adaptively learns the minimal number of patches needed to misclassify the dynamic objects of interest. 
Note that patch-based attacks were used in the contemporary work~\cite{ranjan2019attacking} to attack optical flow models. Here, we study this for semantic segmentation, and introduce an approach to finding the best patch locations, instead of using manually-placed patches as in~\cite{ranjan2019attacking}. 
Furthermore, in contrast to~\cite{ranjan2019attacking}, we study \emph{indirect} attacks that aim to preserve the correct labels within the attacked patch but fool other image regions, and propose detection strategies.
 
When it comes to detecting attacks to semantic segmentation networks, there exist only two techniques~\cite{SC2018,lis2019}. In~\cite{SC2018}, detection is achieved by checking the consistency of predictions obtained from overlapping image patches. In~\cite{lis2019}, the attacked label map is passed through a pix2pix generator~\cite{pix2pix} to re-synthesize an image, which is then compared with the input image to detect the attack. In contrast to these works that need either multiple passes through the network or an auxiliary detector, we detect the attack by analyzing the internal subspaces of the segmentation network. To this end, inspired by the algorithm of~\cite{unifiednips2018} designed for image classification, we compute the Mahalanobis distance of the features to pre-trained class conditional distributions. In contrast to~\cite{SC2018,lis2019}, which study only global image-level detection, we show that our approach is applicable at both the image and the pixel level, yielding the first study on localizing the regions fooled by the attack.


\section{Indirect Local Segmentation Attacks}
\label{sec:attacks}

Let us now introduce our diverse strategies to attack a semantic segmentation network. 
In semantic segmentation, given a clean image $\mathbf{X}\in\mathbb{R}^{W \times H \times C}$,  where $W$, $H$ and $C$ are the width, height, and number of channels, respectively, a  network is trained to minimize a loss function of the form
\begin{equation}\label{eq:semseg}
L({\bf X}) = \sum_{j=1}^{W \times H}{J}({y_j^{true}},  \emph{f}(\mathbf{{X}})_j)\;,
\end{equation}
where \emph{J} is typically taken as the cross-entropy between the true label $y_j^{true}$ and the predicted label $\emph{f}(\mathbf{{X}})_j$ at spatial location $j$. In this context, an adversarial attack is carried out by optimizing for a perturbation that forces the network to output wrong labels for some (or all) of the pixels. Below, we denote by $\bF\in\{\mathbf{0,1}\}^{W \times H}$ the fooling mask such that $\bF_j=1$ if the $j$-th pixel location is targeted by the attacker to be misclassified and $\bF_j=0$ is the predicted label should be preserved. In the remainder of this section, we present our different local attack strategies, and finally introduce our attack detection technique.

\subsection{Indirect Local Attacks}
\label{sec:local}
To study the sensitivity of segmentation networks, we propose to perform local perturbations, confined to predefined regions such as class-specific regions or patches, and to fool other regions than those perturbed.  For example, in the context of automated driving, we may aim to perturb only the regions belonging to road in the input image to fool the car regions in the output label map. This would allow one to modify the physical, static scene while targetting dynamic objects.

Formally, given a clean image  $\mathbf{X}\in\mathbb{R}^{W \times H \times C}$, we aim to find an additive perturbation $\delta\in\mathbb{R}^{W \times H \times C}$ within a perturbation mask $\bM$ that yields erroneous labels within the fooling mask $\bF$.  
To achieve this, we define the perturbation mask $\bM\in\{\mathbf{0,1}\}^{W \times H}$ such that $\bM_j=1$ if the $j$-th pixel location can be perturbed and $\bM_j=0$ otherwise. 

Let ${\bf y}^{pred}_i$ be the label obtained from the clean image at pixel $i$.  An untargeted attack can then be expressed as the solution to the optimization problem
\begin{equation}\label{eq:untargeted}
\begin{aligned}
\delta^* = \arg\min_{\mathbf{\delta}} \sum_{j | \bF_j=1} -\emph{J}(\mathbf{y}_j^{pred}, \emph{f}(\mathbf{{X+M\odot\delta}})_j)\\
+\sum_{j | \bF_j=0}  \emph{J}({\mathbf{y}_j^{pred}}, \emph{f}(\mathbf{{X+M\odot\delta}})_j)\;,
\end{aligned}
\end{equation}
which aims to minimize the probability of ${\bf y}^{pred}_j$ in the targeted regions while maximizing it in the rest of the image.

By contrast, for a targeted attack whose goal is to misclassify any pixel $j$ in the fooling region to pre-defined label $\mathbf{y}^t_j$, we write the optimization problem 
\begin{equation}\label{eq:targeted}
\begin{aligned}
\delta^*=\arg\min_{\mathbf{\delta}} \sum_{j | \bF_j = 1} \emph{J}(\mathbf{y}_j^{t}, \emph{f}(\mathbf{{X+M\odot\delta}})_j) \\
+\sum_{i | \bF_j = 0}  \emph{J}(\mathbf{y}_j^{pred}, \emph{f}(\mathbf{{X+M\odot\delta}})_j)\;.
\end{aligned}
\end{equation}
We solve~\eqref{eq:untargeted} and~\eqref{eq:targeted} using the efficient iterative projected gradient descent algorithm~\cite{pgd2018} with an $\ell_p$-norm perturbation budget $\|\mathbf{\bM \odot \delta}\|_{p}<\epsilon$, where $p\in\{2,\infty\}$.

Note that the formulations above allow one to achieve any local attack. To perform \emph{indirect} local attacks, we can simply define the masks $\bM$ and $\bF$ in such a way that they do not intersect, i.e., $\bM \odot \bF = {\bf 0}$, \kn{ where $\odot$ is element wise product operator}.

\subsection{Adaptive Attacks}
\label{sec:adaptive}
The attacks described in Section~\ref{sec:local} assume the availability of a fixed, predefined perturbation mask $\bM$. In practice, however, one might want to find the best location for an attack, as well as make the attack as local as possible. In this section, we introduce an approach to achieving this by enforcing structured sparsity on the perturbation mask.

To this end, we first re-write the previous attack scheme under an $\ell_2$ budget as an optimization problem. Let $J_t(\bX,\bM,\bF,\delta,\emph{f},\mathbf{y}^{pred},\mathbf{y}^t)$ denote the objective function of either~\eqref{eq:untargeted} or~\eqref{eq:targeted}, where $\mathbf{y}^t$ can be ignored in the untargeted case. Following~\cite{carlini2017towards}, we write an adversarial attack under an $\ell_2$ budget as the solution to the optimization problem
\begin{equation}\label{eq:cwattack}
\delta^* = \arg\min_{\mathbf{\delta}} \lambda_1 \|\mathbf{ \delta}\|_{2}^2  + J_t(\bX,\bM,\bF,\delta, \emph{f},\mathbf{y}^{pred},\mathbf{y}^t)\;,
\end{equation}
where $\lambda_1$ balances the influence of the term aiming to minimize the magnitude of the perturbation.

To identify the best location for an attack together with confining the perturbations to as small an area as possible, we divide the initial perturbation mask $\bM$ into $T$ non-overlapping patches. This can be achieved by defining $T$ masks $\{\bM_t \in\mathbb{R}^{W \times H}\}$ such that, for any $s$, $t$, with $s\neq t$, $\bM_s \odot \bM_t = {\bf 0}$, and $\sum_{t=1}^T \bM_t = \bM$. Our goal then becomes that of finding a perturbation that is non-zero in the smallest number of such masks. This can be achieved by modifying~\eqref{eq:cwattack} as

\begin{equation}\label{eq:GSprior}
\begin{aligned}
\delta^* = \arg\min_{\mathbf{\delta}} \lambda_2 \sum_{t=1}^T \|\mathbf{M}_t\odot \delta\|_{2}  + \lambda_1 \|\mathbf{ \delta}\|_{2}^2 \\
+J_t(\bX,\bM,\bF,\delta, \emph{f}, \mathbf{y}^{pred},\mathbf{y}^t)\;,
\end{aligned}
\end{equation}
whose first term encodes an $l_{2,1}$ group sparsity regularizer encouraging complete groups to go to zero. Such a regularizer has been commonly used in the sparse coding literature~\cite{yuan2006model,nie2010efficient}, and more recently in the context of deep networks for compression purposes~\cite{wen2016learning,alvarez2016learning}. In our context, this regularizer encourages as many as possible of the $\{\mathbf{M_t\odot \delta}\}$ to go to zero, and thus confines the perturbation to a small number of regions that most effectively fool the targeted area $\bF$. $\lambda_2$ balances the influence of this term with the other ones. We then quantify the sparsity of the resulting attack as the percentage of pixels that are perturbed.

\subsection{Universal Local Attacks}
The strategies discussed in Sections~\ref{sec:local} and~\ref{sec:adaptive} are image-specific. To find a universal perturbation effective across all images, we write the optimization problem
\begin{equation}\label{eq:univpatch}
\delta^* = \arg\min_{\mathbf{\delta}} \frac{1}{N}\sum_{i=1}^NJ_u(\bX^i,\bM,\bF^i, \delta, \emph{f}, \kn{\mathbf{y}_{i}^{pred}})
\end{equation}
\MS{Why is there a negative sign?}\KN{ Yes, taking negative out, since its included in the Eq(2) }
where $J_u(\cdot)$ is the objective function for a single image, $N$ is the number of training images, $\mathbf{{X}}^i$ is the $i$-th image \kn{with fooling mask $\bF^i$}, and the mask $\mathbf{M}$ is the global perturbation mask used for all images. In principle, $\mathbf{M}$ can be obtained by sampling patches over all possible image locations. However, we observed such a strategy to be unstable during learning. Hence, in our experiments, we confine ourselves to one or a few \MS{Do we use a few?} \KN{Not reported in any main paper exps, have few numbers that i wanted to report in supplementary}fixed patch positions. Note that, to give the attacker more flexibility, we take the universal attack defined in~\eqref{eq:univpatch} to be an untargeted attack \kn{given in~\eqref{eq:untargeted}.}

\subsection{Adversarial Attack Detection}
\label{sec:detect}

To understand the strength of the attacks discussed above, we introduce a detection method that can act either at the global image level or at the pixel level. The latter is particularly interesting in the case of indirect attacks, where the perturbation regions and the fooled regions are different. In this case, our goal is to localize the pixels that were fooled, which is more challenging that finding those that were perturbed, since their intensity values were not altered.
To this end, we use a score based on the Mahalanobis distance defined on the intermediate feature representations. This is because, as discussed in~\cite{unifiednips2018, lid2018} in the context of image classification, the attacked samples can be better characterized in the representation space than in the output label space.

Specifically, we use a set of training images to compute class-conditional Gaussian distributions, with class-specific means $\mu_{c}^\ell$  and  covariance $\mathbf{\Sigma}^\ell$ shared across \comment{the}\kn{all \emph{C}} \KN{C is referred here in text}classes, 
from the features extracted at \kn{every} intermediate layer $\ell$ of the network within locations corresponding to class label $c$.
We then define a confidence score for each spatial location $j$ in layer $\ell$ as 
\begin{align} \label{eq:mahalanobis}
C(\mathbf{\bX}_{j}^\ell) = \max \limits_{c \in [1,C]}  - \left(\mathbf{\bX}_{j}^\ell - \mathbf{ \mu}_{c}^\ell\right)^\top \mathbf{{\tiny } \Sigma_\ell}^{-1} \left(\mathbf{\bX}_{j}^\ell - \mathbf{ \mu}_{c}^\ell\right)\;,
\end{align}
where $\mathbf{\bX}_{j}^\ell$ denotes the feature vector at location $j$ in layer $\ell$.

We handle the different spatial feature map sizes in different layers by  resizing all of them to a fixed shape.  We then concatenate the confidence scores in all layers at every spatial location and use the resulting $L$-dimensional vectors, with $L$  \kn{being} the number of layers, as input to a  logistic regression classifier with weights $\{\alpha_\ell\}$. We then train this classifier to predict whether a pixel was fooled or not. At test time, we compute the prediction for an image location $j$ as {$\sum_{\ell} \alpha_\ell C(\mathbf{\bX}_{j}^\ell)$}. 

To perform detection at the global image level, we sum over the confidence scores of all spatial positions. That is, for layer $\ell$, we compute an image-level score as $C(\mathbf{\bX}^\ell) =  \sum_j C(\mathbf{\bX}_{j}^\ell)$. We then train another logistic regression classifier using these global confidence scores as input.


\section{Experiments}

\KN{confusion with the term 'attacked' for the readers. Replaced accordingly at below paras}

\label{sec:expresults}
In this section, we 
first explain our experimental setup and implementation details, and then analyze the \comment{degradation} \kn{vulnerability} of state-of-the-art semantic segmentation networks to different types of attacks. Finally, we evaluate our image-level and pixel-level detection strategies.

\noindent{\textbf{Datasets}. In our experiments, we use the Cityscapes~\cite{cityscapesdataset} and Pascal VOC~\cite{pascaldataset} datasets, the two most popular semantic segmentation benchmarks. Specifically, for Cityscapes, we use the complete validation set, consisting of 500 images, for untargeted attacks, but use a subset of 150 images containing dynamic object instances of vehicle classes whose combined area covers at least 8\% of the image for targeted attacks. This is to focus on fooling sufficiently large regions, because reporting results on too small dynamic objects may not be representative of the true behavior of our algorithms.
For Pascal VOC, we use 250 randomly selected images from the validation set because of the limited resources we have access to relative to the large number of experiments we performed. 
	
\noindent\textbf{Models}. We use publicly-available state-of-the-art models, namely FCN~\cite{fcn2015}, \comment{U-Net~\cite{unet2015},}DRNet~\cite{drn2017} , PSPNet~\cite{psp2017}, PSANet~\cite{psanet2018}, DANet~\cite{danet2019} on Cityscapes, and FCN~\cite{fcn2015} and PSANet~\cite{psanet2018} on PASCAL VOC. 
\comment{\ms{Except for U-Net~\cite{unet2015} that we retrained on Cityscapes, we took the official, trained models.}}
FCN, PSANet, PSPNet and DANet share the same ResNet~\cite{he2016deep} backbone network. \comment{, whereas U-Net has a different architecture with encoding and decoding layers.} We perform all experiments at the image resolution of $512 \times 1024$ for Cityscapes and $512 \times 512 $ for PASCAL VOC.  Since different models can have different normalization strategies for the input image, we include normalization in the network and pass the network an input image scaled to [0,1]. More details on the datasets and the models can be found in the supplementary material. 

\noindent\textbf{Adversarial attacks}. We use the iterative projected gradient descent (PGD) method with $\ell_\infty$ and $\ell_2$ norm budgets, as described in Section~\ref{sec:attacks}.  Following~\cite{arnab2018robustness}, we set the number of iterations  for PGD  to  a maximum  of 100,  with an early termination criterion of $90\%$ of attack success rate on the targeted objects. 
\ms{Given the dual objective of the loss functions in~\eqref{eq:untargeted} and~\eqref{eq:targeted}, it may happen that the gradients to maximize the confidence of labels at \comment{non-attacked}  \kn{non-targeted} locations dominate those at \comment{attacked} \kn{targeted} ones. Hence, as suggested in~\cite{univperturb}, we ignore the loss at locations where the label is predicted correctly as the target label with a confidence of at least 0.3.}
We evaluate $\ell_\infty$ attacks with a step size $\alpha \in \{1\text{e-}5, 1\text{e-}4,1\text{e-}3,5\text{e-}3\}$. For $\ell_2$ attacks, we set $\alpha \in \{8\text{e-}3, 4\text{e-}2,8\text{e-}2\}$.  We perform two types of attacks; targeted and untargeted. The untargeted attacks focus on fooling the network to move away from the {predicted label.} For the targeted attacks, we chose a safety-sensitive goal, and thus aim to fool the dynamic object regions to be misclassified as their (spatially) nearest background label. We do not use ground-truth information in any of the experiments but perform attacks based on the predicted labels only. We implement our algorithms in PyTorch~\cite{paszke2017automatic} using the advertorch library~\cite{ding2019advertorch} on a single Tesla 32GB GPU.

\noindent\textbf{Evaluation metric}.  Following~\cite{univperturb, arnab2018robustness, DAG2017}, we report the mean Intersection over Union (mIoU) and  Attack Success Rate  (ASR) computed over the entire dataset.  The mIoU of  FCN~\cite{fcn2015}, \comment{U-Net~\cite{unet2015},} DRNet~\cite{drn2017}, PSPNet~\cite{psp2017}, PSANet~\cite{psanet2018}, and DANet~\cite{danet2019}  on clean samples at full resolution are 0.66, \comment{0.39,} 0.64, 0.73, 0.72, 0.67, respectively. For targeted attacks, we report the average $\text{ASR}_{t}$, computed as the percentage of pixels that were predicted as the target label. We additionally report the $\text{mIoU}_{u}$, which is computed between the adversarial and normal sample predictions. 
For untargeted attacks, we report the  $\text{ASR}_{u}$,  computed as the percentage of pixels that were assigned to a different class than their normal label prediction. 
Since, in most of our experiments, the fooling region is confined to local objects, we compute the above metrics only at the fooling mask regions. We observed that the \comment{non-attacked}  \kn{non-targeted} regions retain their prediction label more than 98\% of the time, and hence we report the metrics at \comment{non-attacked}  \kn{non-targeted} regions in the supplementary material.
To evaluate the detection of adversarial attacks, we  report the Area under the Receiver Operating Characteristics (AUROC), both at image level, as in~\cite{SC2018,lis2019}, and at pixel level.

\subsection{Indirect Attacks}\label{sec:expindirectattack}

Let us study the sensitivity of the networks to indirect local attacks. In this setting, we first perform a  targeted  attack, formalized in~\eqref{eq:targeted}, to fool the dynamic object areas by allowing the attacker to perturb any region belonging to the static object classes. This is achieved by setting the perturbation mask $\bM$ to 1 at all the static class pixels and the fooling mask $\bF$ to 1 at all the dynamic class pixels.
We 
report the $\text{mIoU}_{u}$  and  $\text{ASR}_{t}$ metrics in Table~\ref{tab:context_perturb_city_linf} and~\ref{tab:context_perturb_city_l2} \MS{What do the bold numbers mean in the tables?} \KN{most vulnerable model, not sure where should i specify that} on Cityscapes for $\ell_\infty$ and $\ell_2$ attacks, respectively. As evidenced from the tables, FCN is more robust to such indirect attacks than the networks that leverage contextual information. In particular, PSANet  \comment{, which uses long range contextual dependencies,} and PSPNet  are highly sensitive to these attacks.


\begin{table}
	\begin{subtable}[t]{\linewidth}
		\centering
		{\small
			\resizebox{\columnwidth}{!}{%
				\begin{tabular}{lccccc}
									\midrule
					\multirow{1}{*}{Network}  & Attack & $\alpha = 0.00001$ &   $ \alpha =0.0001 $ &  $\alpha=0.001$ &  $\alpha=0.005$    \\ 
					\midrule
					FCN~\cite{fcn2015} &   \multirow{6}{*}{$\ell_\infty$} & 0.64 / 5.0\% &0.28 / 29\%& 0.13 / 55\%& 0.11 / 61\%  \\ 
					PSPNet~\cite{psp2017} &&0.70 / 12\% & 0.05 / 85\% & 0.00 / 89\% & 0.00 / 90\%\\
					PSANet~\cite{psanet2018} &&{ \bf0.59 / 14\%} & {  0.03 / 85\%} &{  0.01 / 90\% }& {\bf 0.00 / 90\%} \\
					DANet~\cite{danet2019} & & 0.80 / 5.0\%  & 0.11 / 79\% & 0.01 / 90\%& 0.00 / 90\%\\
					DRN~\cite{drn2017} && 0.64 / 6.0\%& 0.15 / 56\% & 0.03 / 84\% & 0.02 / 86\% \\
					\hline
				\end{tabular}
			}
		}
		\caption{$\ell_\infty$ attack } 
		\label{tab:context_perturb_city_linf}
		\vspace{0.2cm}
	\end{subtable}\hfill

	\begin{subtable}[t]{\linewidth}
		\centering
		{\small
			\resizebox{0.8\columnwidth}{!}{%
			\begin{tabular}{lcccc}
								\midrule
				\multirow{1}{*}{Network}  & Attack  &   $ \alpha =0.008 $ &  $\alpha=0.04$ & $\alpha=0.08$    \\ 
				\midrule
				FCN~\cite{fcn2015} &   \multirow{6}{*}{$\ell_2$} & 0.60 / 10\% & 0.56 / 26\%& 0.27 / 36\%  \\ 
				PSPNet~\cite{psp2017} &&{ \bf 0.67 / 19\%}& { \bf 0.23 / 67\%} & { \bf 0.06 /  84\%} \\
				PSANet~\cite{psanet2018} && { 0.59 / 14\%} & {  0.21 / 63\%}& { 0.06 / 82\%}\\ 
				DANet~\cite{danet2019} &&0.79 / 11\% & 0.43 / 49\% & 0.13 / 79\%\\
				DRN~\cite{drn2017} && 0.63 /  10\%& 0.24 / 47\% & 0.13 / 64\% \\
				\hline
			\end{tabular}%
			}
		}
		\caption{ $\ell_2$ attack } 
		\label{tab:context_perturb_city_l2}
	\end{subtable}\hfill
	\caption{{\small {\bf Indirect Attacks} on Cityscapes  to fool dynamic classes while perturbing static ones. The numbers indicate $\text{mIoU}_{u}$/$\text{ASR}_{t}$, obtained using different step sizes $\alpha$ for $\ell_\infty$  and $\ell_2$ attacks. } }
	\label{tbl:main}
\end{table}
 
To further understand the impact of indirect \emph{local} attacks, we constrain the perturbation region to a subset of the static class regions. To do this in a systematic manner, we  perturb the static class regions that are at least $d$ pixels away from any dynamic object, and vary the value $d$. The results of this experiment using $\ell_2$ and $\ell_\infty$ attacks are provided in Table~\ref{tab:boundary_city}. Here, we chose a step size $\alpha = 0.005$  for $\ell_\infty$ and $\alpha = 0.08$  for $\ell_2$. Similar conclusions as in the previous non-local scenario can be drawn: Modern networks that use larger receptive fields are extremely vulnerable to such perturbations, even when they are far away from the targeted regions. By contrast, FCN is again more robust. For example, as shown in Figure~\ref{fig:boundary_city}, while an adversarial attack occurring 100 pixels away from the nearest dynamic objects has a high success rate on the context-aware networks, the FCN predictions remain accurate.

\begin{table}[t]
	\centering
	{\footnotesize 
	\resizebox{0.95\columnwidth}{!}{%
		\begin{tabular}{lccccc}
			\hline
			\multirow{1}{*}{Network}  & Attack & $d = 0$ &   $d =50$ &   $ d =100 $ &  $\text{d}=150$    \\ 
						\midrule
			  FCN~\cite{fcn2015} &   \multirow{6}{*}{$\ell_\infty$} & 0.11 / 64\%& 0.77 / 2.0\% &0.98 / 0\%&1.00 / 0.0\% \\ 
			 PSPNet~\cite{psp2017} & & 0.00 / 90\%   & 0.14 / 73\% &0.24 / 60\%& 0.55 / 23\%    \\
			 PSANet~\cite{psanet2018} &&  {  0.00 / 90\% }  &{  {\bf0.11} / 71\%}&{ \bf 0.13 / 65\%} &{\bf 0.29 / 47\% }\\
			 DANet~\cite{danet2019} &&   0.00 / 90\%  & 0.13 / { \bf 81\%}&0.48 / 43\%& 0.80 / 10\% \\
			 DRN~\cite{drn2017} &    & 0.02 / 86\%& 0.38 /  22\%& 0.73 / 3\%& 0.94 / 1.0\% \\
			 \midrule			
			FCN~\cite{fcn2015}  &   \multirow{6}{*}{$\ell_2$}& 0.27 / 36\% &0.79 / 2.0\%&   0.98 / 2.0\% & 0.99 / 1.0\% \\ 
			PSPNet~\cite{psp2017} &    &  0.06 / {  \bf 84\%}& { 0.18 /  73\%}&  {  0.55 / 23\% } & { 0.99 / 0.0\%}\\
			PSANet~\cite{psanet2018} &   & {  0.06 / 82\%}&{ \bf 0.10 / 75\%}&{\bf  0.14 / 66 \%}&  {\bf  0.31 / 44\% }\\
			DANet~\cite{danet2019} &    & 0.13 / 79\% & 0.27 / 71\%&0.67 / 26\% & 0.85 / 7.0\% \\
			DRN~\cite{drn2017} &    & 0.13 / 64\% & 0.44 / 17\%& 0.76 / 3.0\%& 0.95 / 0.0\%\\
			\hline
		\end{tabular}%
	}
}
	\caption{{\bf Impact of Local Attacks} by perturbing pixels that are at least $d$ pixels away from any dynamic class. We report $\text{mIoU}_{u}$/$\text{ASR}_{t}$ for different values of $d$. } 
	\label{tab:boundary_city}
\end{table}


\providecommand{\localwidth}{}
\renewcommand{\localwidth}{\linewidth}
\providecommand{\localheight}{}
\renewcommand{\localheight}{1.5cm}

\begin{figure}[t]
	\centering\offinterlineskip
	\centering
	\includegraphics[width=\localwidth]{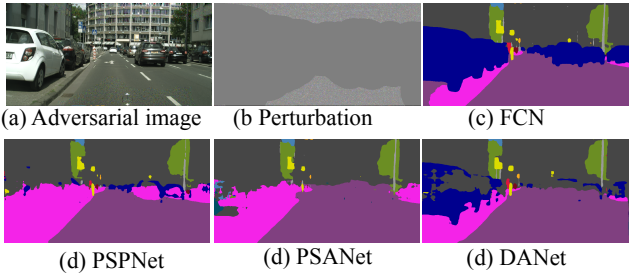}
	\caption{{\bf Indirect Local attack} on different networks with perturbations at least $d = 100$ pixels away from any dynamic class. 
} 
	\label{fig:boundary_city} 
	\vspace*{-5mm}
\end{figure}

\subsection{Adaptive Indirect Local Attacks}
We now study the impact of our approach to adaptively finding the most sensitive context region to fool the dynamic objects.  To this end,
we use the group sparsity based optimization given in~\eqref{eq:GSprior} and find the minimal perturbation region to fool all dynamic objects to their nearest static label. Specifically, we achieve this in two steps. First, we divide the perturbation mask $\bM$ corresponding to all static class pixels into uniform patches of size $h \times w $, and find the most sensitive ones by solving~\eqref{eq:GSprior} with a relatively large group sparsity weight $\lambda_{2}=100.0 $. Second, we limit the perturbation region by selecting the $n$ patches that have the largest values  $\|\mathbf{M}_t\odot \delta\|_{2}$), choosing $n$ so as to achieve a given sparsity level $S\in\{75\%,85\%,90\%,95\%\}$. Specifically, $S$ is computed as the percentage of perturbed pixels relative to the initial perturbation mask.  We then re-optimize~\eqref{eq:GSprior} with $\lambda_{2}=0.0 $. In both steps, we set $\lambda_1=0.01$ and use the Adam optimizer~\cite{kingma2014adam} with a learning rate  of $0.01$ and a patch size $h=60$, $w=120$. We clip the perturbation values below 0.005 to 0.0 at each iteration.  This results in very local perturbation regions, active only in the most sensitive areas, as shown in Figure~\ref{fig:adaptive_city} for PSANet on Cityscapes. As shown in Table~\ref{tab:adaptivepatch_city}, all context-aware networks are significantly affected by such perturbations, even when they are confined to small background regions. This means that, in the physical world, an attacker could add a small sticker at a static position to essentially make dynamic objects disappear from the network's view.

For PASCAL VOC, we use the same hyperparameter values except for $\lambda_{2}$, which is set to 10.0 in the first optimization step. Furthermore, we set the patch size to $h=60$, $w=60$.  As shown in Figure~\ref{fig:adaptive_voc}, we are able to find the most sensitive regions that cover a minimum area in the static class to fool the dynamic foreground objects. We report the effect of our adaptive indirect attacks on PSANet and FCN in Table~\ref{tab:adaptivepatch_voc}.\kn{ For instance, at high sparsity level of $95\%$, PSANet has  $\text{ASR}_{t}$ of $44\%$ compared to $1\%$ for FCN.}


\begin{table}[t]
	\centering
	{\small
		\resizebox{0.95\columnwidth}{!}{%
			\begin{tabular}{lcccc}
				\midrule
				\multirow{1}{*}{Network}   & $S=75\%$ &  $S=85\%$ &  $S=90\%$ & $S=95\%$     \\ 
				\midrule
				\multirow{1}{*}{FCN~\cite{fcn2015}} & 0.52 / $12\%$ & 0.66 / $6\%$ & 0.73 / $4\%$ & 0.84 / 1.0\% \\
				\multirow{1}{*}{PSPNett~\cite{psp2017}}   & 0.19 / 70\%   & 0.31 / 54\% & 0.41 / 42\%&  0.53 / 21\% \\
				\multirow{1}{*}{PSANet~\cite{psanet2018}} & \textbf{0.10 / 78\%} & \textbf{0.16 / 71\%} & \textbf{0.20 / 64\%} & \textbf{0.35 / 44\%}\\
				\multirow{1}{*}{DANet~\cite{danet2019}} & 0.30 / 64\%& 0.52 / 43\% & 0.64 / 30\%&  0.71 / 21\% \\
				\multirow{1}{*}{DRN~\cite{drn2017}} & 0.42 / 23\%& 0.55 / 13\% & 0.63 / 9.0\%&  0.77 / 4.5\% \\
				\hline
			\end{tabular}%
		}
	}
	\caption{{{\small \bf Adaptive Indirect Local Attacks} on Cityscapes. We report the $\text{mIoU}_{u}$/$\text{ASR}_{t}$ for different sparsity levels $S$}.}
	\label{tab:adaptivepatch_city}
\end{table}

\comment{
\begin{table}[t]
	\centering
	{\normalsize
	\resizebox{0.95\columnwidth}{!}{%
		\begin{tabular}{|c|c|c|c|c|c|}
			\hline
			\multirow{1}{*}{Network}  & Metrics & $S=95\%$ &  $S=90\%$ &  $S=85\%$ & $S=75\%$     \\ 
			\hline
			\multirow{2}{*}{FCNet~\cite{psp2017}} &  $\ell_2$-norm && & & \\
			&  mIoU / mSR&&&&\\
		 \midrule
		 \multirow{2}{*}{PSPNet~\cite{psp2017}} &  $\ell_2$-norm && &&  \\
		 &  mIoU / mSR&&&&\\
		 \midrule
		 \multirow{2}{*}{PSANet~\cite{psp2017}} &  $\ell_2$-norm &&& &  \\
		 &  mIoU / mSR&&&&\\
		 \midrule
		 \multirow{2}{*}{DANet~\cite{psp2017}} &  $\ell_2$-norm && &&  \\
		 &  mIoU / mSR&&&&\\
		 \midrule
		\end{tabular}%
	}
}
	\caption{{\bf Adaptive Indirect Local Attacks} on cityscapes dataset. }
	\label{tab:adaptivepatch_city}
	\vspace*{-0.3cm}
\end{table}
}

\providecommand{\localwidth}{}
\renewcommand{\localwidth}{0.20\linewidth}

\begin{figure*}[t]
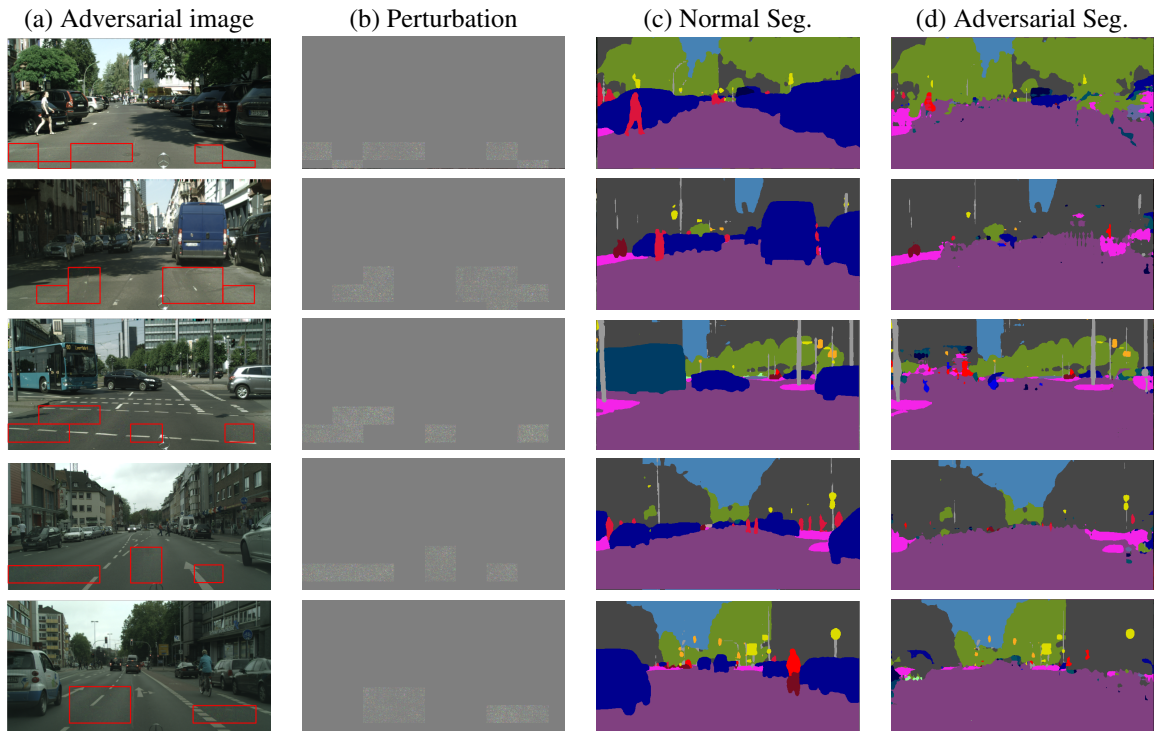

	\centering
	
	\begin{tabular}{cccc}
		(a) Adversarial image& (b) Perturbation & (c) Normal Seg.& (d) Adversarial Seg.\\
		

	\includegraphics[width=\localwidth]{\expfigadaptivecity{165_image1.png}}&
\includegraphics[width=\localwidth]{\expfigadaptivecity{165_perturb.png}}&
\includegraphics[width=\localwidth]{\expfigadaptivecity{165_realseg.png}}&
\includegraphics[width=\localwidth]{\expfigadaptivecity{165_advseg.png}}
\\	

			\includegraphics[width=\localwidth]{\expfigadaptivecity{130_image1.png}}&
		\includegraphics[width=\localwidth]{\expfigadaptivecity{130_perturb.png}}&
		\includegraphics[width=\localwidth]{\expfigadaptivecity{130_realseg.png}}&
		\includegraphics[width=\localwidth]{\expfigadaptivecity{130_advseg.png}}
		\\	
		
		\includegraphics[width=\localwidth]{\expfigadaptivecity{145_image1.png}}&
		\includegraphics[width=\localwidth]{\expfigadaptivecity{145_perturb.png}}&
		\includegraphics[width=\localwidth]{\expfigadaptivecity{145_realseg.png}}&
		\includegraphics[width=\localwidth]{\expfigadaptivecity{145_advseg.png}}
		\\	
			\includegraphics[width=\localwidth]{\expfigadaptivecity{340_image1.png}}&
		\includegraphics[width=\localwidth]{\expfigadaptivecity{340_perturb.png}}&
		\includegraphics[width=\localwidth]{\expfigadaptivecity{340_realseg.png}}&
		\includegraphics[width=\localwidth]{\expfigadaptivecity{340_advseg.png}}
		\\	
			\includegraphics[width=\localwidth]{\expfigadaptivecity{395_image1.png}}&
		\includegraphics[width=\localwidth]{\expfigadaptivecity{395_perturb.png}}&
		\includegraphics[width=\localwidth]{\expfigadaptivecity{395_realseg.png}}&
		\includegraphics[width=\localwidth]{\expfigadaptivecity{395_advseg.png}}
		\\	
	\end{tabular}
	
	\vspace{-3mm}
	\caption{{\bf Adaptive Indirect Local Attacks on Cityscapes with PSANet~\cite{psanet2018}}. An adversarial input image {\bf(a)} when attacked at positions shown as red boxes with a perturbation {\bf(b)} is \comment{mis-classifies the dynamic object classes in the normal segmentation map}   \kn{mis-classified at dynamic object areas in the normal segmentation map {\bf(c)} to result in  {\bf(d)}. }
	}
	\label{fig:adaptive_city}
\end{figure*}


\providecommand{\localwidth}{}
\renewcommand{\localwidth}{\linewidth}

\providecommand{\localheight}{}
\renewcommand{\localheight}{2cm}

\begin{figure*}[hbt!]
	\centering
	\includegraphics[width=\localwidth]{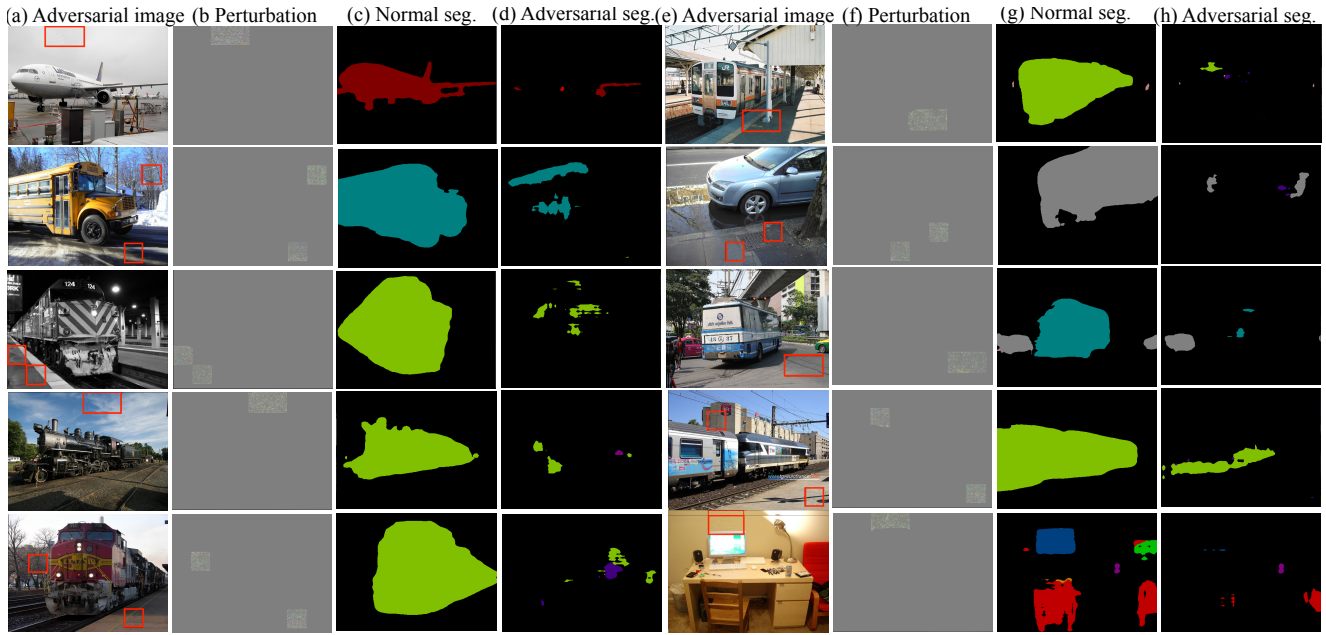}
	\caption{{\bf Adaptive Indirect Local Attacks on PASCAL VOC with PSANet~\cite{psanet2018}}. An adversarial input image {\bf(a),(e)} when attacked at positions shown as red boxes with a perturbation {\bf(b),(f)}
	\comment{mis-classifies the foreground object classes in the normal segmentation map in}
	\kn{is mis-classified at foreground object areas in the normal segmentation map in} {\bf(c), (g)} to result in  {\bf(d), (h)}, respectively. }
	\label{fig:adaptive_voc}
\end{figure*}


\begin{table}[t]
	\centering
	{\small
		\resizebox{0.95\columnwidth}{!}{%
			\begin{tabular}{lcccc}
				\midrule
				\multirow{1}{*}{Network}   & $S=75\%$ &  $S=85\%$ &  $S=90\%$ & $S=95\%$     \\ 
				\midrule
				\multirow{1}{*}{FCN~\cite{fcn2015}} & 0.52 / $12\%$ & 0.66 / $6.0\%$ & 0.73 / $4.0\%$ & 0.84 / 1.0\% \\
				\multirow{1}{*}{PSANet~\cite{psanet2018}} & \textbf{0.10 / 78\%} & \textbf{0.16 / 71\%} & \textbf{0.20 / 64\%} & \textbf{0.35 / 44\%}\\
				\hline
			\end{tabular}%
		}
	}
	\caption{{{\small \bf Adaptive Indirect Local Attacks} on PASCAL VOC. We report the $\text{mIoU}_{u}$/$\text{ASR}_{t}$ for different  sparsity levels $S$}.}
	\label{tab:adaptivepatch_voc}
\end{table}

\comment{
\begin{table}[t]
	\centering
	{\normalsize
	\resizebox{0.95\columnwidth}{!}{%
		\begin{tabular}{|c|c|c|c|c|c|}
			\hline
			\multirow{1}{*}{Network}  & Metrics & $S=95\%$ &  $S=90\%$ &  $S=85\%$ & $S=75\%$     \\ 
			\hline
			\multirow{2}{*}{FCNet~\cite{psp2017}} &  $\ell_2$-norm && & & \\
			&  mIoU / mSR&&&&\\
		 \midrule
		 \multirow{2}{*}{PSPNet~\cite{psp2017}} &  $\ell_2$-norm && &&  \\
		 &  mIoU / mSR&&&&\\
		 \midrule
		 \multirow{2}{*}{PSANet~\cite{psp2017}} &  $\ell_2$-norm &&& &  \\
		 &  mIoU / mSR&&&&\\
		 \midrule
		 \multirow{2}{*}{DANet~\cite{psp2017}} &  $\ell_2$-norm && &&  \\
		 &  mIoU / mSR&&&&\\
		 \midrule
		\end{tabular}%
	}
}
	\caption{{\bf Adaptive Indirect Local Attacks} on cityscapes dataset. }
	\label{tab:adaptivepatch_city}
	\vspace*{-0.3cm}
\end{table}
}


\begin{table}[t]
		\large
	\centering
	{

		\resizebox{0.99\columnwidth}{!}{%
			\begin{tabular}{lcccc}
				\midrule
				\multirow{1}{*}{Network}   & $51 \times 102 \mathbf{ (1.0\%)}$ &  $76 \times 157 \mathbf{ (2.3\%)}$ &  $102 \times 204 \mathbf{ (4.0\%)}$&  $153\times 306 \mathbf{ (9.0\%)}$  \\ 
				\midrule
				FCN~\cite{fcn2015}  & 0.85 / 2.0\%& 0.78 / 4.0\% & 0.73 / 9.0\% &  0.58 / 18\%\\ 
				PSPNet~\cite{psp2017}  &0.79 / 3.0\%& 0.63 / 11\%& 0.44 / 27\% & 0.08 / 83\%  \\
				PSANet~\cite{psanet2018}  &\textbf{0.41} / \textbf{37\%}& \textbf{0.22} / \textbf{60\%}   & \textbf{0.14} / \textbf{70\%} & \textbf{0.10} / \textbf{90\%}\\
				DANet~\cite{danet2019}  &0.79 / 4.0\% & 0.71 / 10\% & 0.65 / 15\%& 0.40 / 42\%\\
				DRN~\cite{drn2017} & 0.82 / 3.0\%& 0.78 /  8.0\% &  0.71 / 14\%& 0.55 / 28\%  \\
				\hline
			\end{tabular}%
		}
	}
\caption{{\small\textbf{Universal Local Attacks}. We show the impact of patch size  \kn{$h$$\times$$w$  (\text{area}\%)} on different networks and report  $\text{mIoU}_{u}$/$\text{ASR}_{u}$.}}
\label{tab:univpatch_city}
	\vspace*{-0.1cm}
\end{table}

\subsection{Universal Local Attacks}\label{sec:expuniv}

\providecommand{\localwidth}{}
\renewcommand{\localwidth}{0.18\linewidth}

\providecommand{\localheight}{}
\renewcommand{\localheight}{2cm}

\begin{figure*}[htp]
	\centering
	\begin{tabular}{ccccc}
		(a) Adversarial image& (b) FCN~\cite{fcn2015} & (c) PSPNet~\cite{psp2017}& (d) PSANet~\cite{psanet2018} & (e) DANet~\cite{danet2019}\\ 
		\includegraphics[height= \localheight, width=\localwidth]{\expfigunivcity{201_image.png}}&
		\includegraphics[height= \localheight,width=\localwidth]{\expfigunivcity{201_fcn.png}}&
			\includegraphics[height= \localheight,width=\localwidth]{\expfigunivcity{201_psp.png}}&
				\includegraphics[height= \localheight,width=\localwidth]{\expfigunivcity{201_psa.png}}&
					\includegraphics[height= \localheight,width=\localwidth]{\expfigunivcity{201_danet.png}}\\
					\end{tabular}
	
	\vspace{-3mm}
	\caption{{\bf  Universal Local Attacks} on segmentation networks. The degradation in FCN~\cite{fcn2015} is limited to the attacked area, whereas for context-aware networks, such as PSPNet~\cite{psp2017}, PSANet~\cite{psanet2018}, DANet~\cite{danet2019}, it extends to far away regions.}
	\label{fig:univ_city}
\end{figure*}

In this section, instead of considering image-dependent perturbations, we study the existence of universal local perturbations and their impact on semantic segmentation networks. In this setting, we perform untargeted \emph{local} attacks by placing a fixed-size patch at a predetermined position. While the patch location can in principle be sampled at any location, we found learning its position to be unstable to due to the large number of possible patch locations in the entire dataset. Hence, here, we consider the scenario where the patch is located at the center of the image. We then learn a local perturbation that can fool the entire dataset of images for a given network by optimizing the objective given in~\eqref{eq:univpatch}.  Specifically, the perturbation mask $\bM$ is active only at the  patch location and the fooling mask $\bF$ at all image positions, i.e., at both static and dynamic classes. We learn the universal local perturbation using $100$ images from Cityscapes and use the remaining $400$ images for evaluation purpose. 
We use $\ell_{\infty}$ optimization with $\alpha = 0.001$ for 200 epochs on the training set.
We report the results of such universal patch attacks in Table~\ref{tab:univpatch_city} for different patch sizes. 
As shown in the table, PSANet and PSPNet are vulnerable to such universal attacks, even when only $2.3\%$ of the image area is perturbed. From Figure~\ref{fig:univ_city}, we can see that the fooling region propagates to a large area far away from the perturbed one.
{\tiny }


\subsection{Attack Detection}\label{sec:attackdetectionexps}



\begin{table}[h]
\begin{center}
	{ 
	\resizebox{\linewidth}{!}{%
			\begin{tabular}{lcccccc}
				\hline
				\multirow{2}{*}{Networks}   & Perturbation  & Fooling  & {$\ell_{\infty}$ / $\ell_2$} & $\text{Mis.}$ & Global AUROC & Local AUROC  \\  
				&   region& region &norm & pixels $\%$& { SC~\cite{SC2018} / Re-Syn~\cite{lis2019} / \ours} &\ours \\
				\midrule
				\multirow{4}{*}{FCN \cite{fcn2015}} &
				 Global & \full        & 0.10 / 17.60 & $90 \%$ & \textbf{1.00 / 1.00} / {0.94}& 0.90  \\
				&  \up & \full        & 0.30 / 37.60& $4 \%$ & 0.71 / 0.63 / \textbf{1.00}&0.94  \\
				&     \fs   &  \dyn   & 0.07 / 2.58 & $13\%$& 0.57 / 0.71 / \textbf{1.00}& 0.87 \\
				&  \ap & \dyn  & 0.14 / 3.11& $1.7\%$&0.51 / {0.65} / \textbf{0.87} & 0.89 \\
			\midrule
\multirow{4}{*}{PSPNet \cite{psp2017}}& 
				 Global & \full        & 0.06 / 10.74 & $83\%$ & 0.90 / \textbf{1.00} /  {0.99}& 0.85  \\
				 & \up & \full        & 0.30 / 38.43&$11\%$&0.66 / 0.70 /  \textbf{1.00}& 0.96  \\
&     \fs   &  \dyn   & 0.03 / 1.78 & $14\%$& 0.57 / 0.75 / \textbf{0.90} & 0.87 \\
&  \ap & \dyn  & 0.11 / 5.25& $11\%$& 0.57 / 0.75 / \textbf{0.90}& 0.82 \\
\midrule
\multirow{4}{*}{PSANet \cite{psanet2018}} &
				 Global & \full        & 0.05 / 8.26 & $92 \%$ & 0.90 / 1.00 / {1.00}& 0.67  \\
				 &  \up & \full        &0.30 / 38.6 & $60\%$& 0.65 / 1.00 / 1.00 & 0.98 \\
&     \fs   &  \dyn   & 0.02 / 1.14  &$12\%$& 0.61 / 0.76 / \textbf{1.00} &  0.92\\
&  \ap & \dyn  & 0.10 / 5.10&$10\%$&0.50 / 0.82 / \textbf{1.00}& 0.94 \\
\midrule
\multirow{4}{*}{DANet \cite{danet2019}} &
				 Global & \full        & 0.06 / 12.55& $82 \%$ & 0.89 / 1.00 / {1.00}& 0.68  \\
				 &  \up & \full        & 0.30 / 37.20 & $10\%$&0.67 / {0.63} / \textbf{0.92}& 0.89 \\
&     \fs   &  \dyn   & 0.05 / 1.94 &$13\%$&0.57 / 0.69 / \textbf{0.94} & 0.88 \\
&  \ap & \dyn  & 0.14 / 6.12 &$43\%$&0.59 / 0.68 / \textbf{0.98}& 0.82 \\
\midrule				
			\end{tabular}
		}
	}
\end{center}
\vspace*{-0.4cm}
\caption{ {\small \textbf{Attack detection} on Cityscapes with different perturbation settings.}}
\label{tbl:attack_detection}
\end{table}

We now turn to studying the effectiveness of the attack detection strategies described in Section~\ref{sec:detect}. We also compare our approach to the only two detection techniques that have been proposed for semantic segmentation~\cite{SC2018,lis2019}. The method in~\cite{SC2018} uses the spatial consistency of the predictions obtained from $K=50$ random overlapping patches of size $256 \times 256$. The one in~\cite{lis2019} compares an image re-synthesized from the predicted labels with the input image. Both methods were designed to handle attacks that fools the entire label map, unlike our work where we aim to fool local regions. Furthermore, both methods perform detection at the image level, and thus, in contrast to ours, do not localize the fooled regions at the pixel level.

We study detection in \comment{three} four perturbation settings: Global image perturbations (Global) to fool the entire image;  Universal patch perturbations (UP) at a fixed location to fool the entire image; Full static (FS) class perturbations to fool the dynamic classes; Adaptive patch (AP) perturbations in the static class regions to fool the dynamic objects. As shown in Table~\ref{tbl:attack_detection}, while the state-of-the-art methods~\cite{SC2018,lis2019}  \comment{work well} \kn{has high Global AUROC}\KN{Refer teerms Global AUROC and Local AUROC in text} in the first setting where the entire image is targeted, our detection strategy outperforms them by a large margin in the other scenarios. We believe this to be due to the fact that, with local attacks, the statistics obtained by studying the consistency across local patches, as in~\cite{SC2018}, are much closer to the clean image statistics. Similarly, the image re-synthesized by a pix2pix generator, as used in~\cite{lis2019}, will look much more similar to the input one in the presence of local attacks instead of global ones. For all the perturbation settings, we also report the mean percentage of pixels misclassified relative to the number of pixels in the image. We provide additional detection results with different perturbation settings and noise levels in the supplementary material.



\section{Conclusion}
\label{sec:conclusion}
In this paper, we have studied the impact of indirect local image perturbations on the performance of modern semantic segmentation networks.
We have observed that the state-of-the-art segmentation networks, such as PSANet and PSPNet, are more vulnerable to local perturbations because their use of context, which improves their accuracy on clean images, enables the perturbations to be propagated to distant image regions. 
As such, they can be attacked by perturbations that cover as little as $2.3\%$ of the image area.  We have then proposed a Mahalanobis distance-based detection strategy which has proven effective at image-level attack detection. While promising, its performance at localizing the fooled regions in a pixel-wise manner still leaves room for improvement, and addressing this will be our goal in the future.\\

{\small
\bibliographystyle{ieee_fullname}
\bibliography{egbib}

\begin{thebibliography}{10}\itemsep=-1pt

\bibitem{alvarez2016learning}
Jose~M Alvarez and Mathieu Salzmann.
\newblock Learning the number of neurons in deep networks.
\newblock In {\em Advances in Neural Information Processing Systems}, pages
  2270--2278, 2016.

\bibitem{arnab2018robustness}
Anurag Arnab, Ondrej Miksik, and Philip~HS Torr.
\newblock On the robustness of semantic segmentation models to adversarial
  attacks.
\newblock In {\em Proceedings of the IEEE Conference on Computer Vision and
  Pattern Recognition}, pages 888--897, 2018.

\bibitem{pgd2018}
Anish Athalye, Nicholas Carlini, and David Wagner.
\newblock Obfuscated gradients give a false sense of security: Circumventing
  defenses to adversarial examples.
\newblock {\em arXiv preprint arXiv:1802.00420}, 2018.

\bibitem{carlini2017towards}
Nicholas Carlini and David Wagner.
\newblock Towards evaluating the robustness of neural networks.
\newblock In {\em 2017 IEEE Symposium on Security and Privacy (SP)}, pages
  39--57. IEEE, 2017.

\bibitem{cityscapesdataset}
Marius Cordts, Mohamed Omran, Sebastian Ramos, Timo Rehfeld, Markus Enzweiler,
  Rodrigo Benenson, Uwe Franke, Stefan Roth, and Bernt Schiele.
\newblock The cityscapes dataset for semantic urban scene understanding.
\newblock In {\em Proceedings of the IEEE conference on computer vision and
  pattern recognition}, pages 3213--3223, 2016.

\bibitem{ding2019advertorch}
Gavin~Weiguang Ding, Luyu Wang, and Xiaomeng Jin.
\newblock Advertorch v0. 1: An adversarial robustness toolbox based on pytorch.
\newblock {\em arXiv preprint arXiv:1902.07623}, 2019.

\bibitem{dong2018boosting}
Yinpeng Dong, Fangzhou Liao, Tianyu Pang, Hang Su, Jun Zhu, Xiaolin Hu, and
  Jianguo Li.
\newblock Boosting adversarial attacks with momentum.
\newblock In {\em Proceedings of the IEEE conference on computer vision and
  pattern recognition}, pages 9185--9193, 2018.

\bibitem{pascaldataset}
Mark Everingham, Luc Van~Gool, Christopher~KI Williams, John Winn, and Andrew
  Zisserman.
\newblock The pascal visual object classes challenge 2007 (voc2007) results.
\newblock 2007.

\bibitem{robustphysical2018}
Kevin Eykholt, Ivan Evtimov, Earlence Fernandes, Bo Li, Amir Rahmati, Chaowei
  Xiao, Atul Prakash, Tadayoshi Kohno, and Dawn Song.
\newblock Robust physical-world attacks on deep learning visual classification.
\newblock In {\em Proceedings of the IEEE Conference on Computer Vision and
  Pattern Recognition}, pages 1625--1634, 2018.

\bibitem{danet2019}
Jun Fu, Jing Liu, Haijie Tian, Yong Li, Yongjun Bao, Zhiwei Fang, and Hanqing
  Lu.
\newblock Dual attention network for scene segmentation.
\newblock In {\em Proceedings of the IEEE Conference on Computer Vision and
  Pattern Recognition}, pages 3146--3154, 2019.

\bibitem{gonfaus2010harmony}
Josep~M Gonfaus, Xavier Boix, Joost Van~de Weijer, Andrew~D Bagdanov, Joan
  Serrat, and Jordi Gonzalez.
\newblock Harmony potentials for joint classification and segmentation.
\newblock In {\em 2010 IEEE computer society conference on computer vision and
  pattern recognition}, pages 3280--3287. IEEE, 2010.

\bibitem{goodfellow2014explaining}
Ian~J Goodfellow, Jonathon Shlens, and Christian Szegedy.
\newblock Explaining and harnessing adversarial examples.
\newblock {\em arXiv preprint arXiv:1412.6572}, 2014.

\bibitem{hariharan2015hypercolumns}
Bharath Hariharan, Pablo Arbel{\'a}ez, Ross Girshick, and Jitendra Malik.
\newblock Hypercolumns for object segmentation and fine-grained localization.
\newblock In {\em Proceedings of the IEEE conference on computer vision and
  pattern recognition}, pages 447--456, 2015.

\bibitem{he2016deep}
Kaiming He, Xiangyu Zhang, Shaoqing Ren, and Jian Sun.
\newblock Deep residual learning for image recognition.
\newblock In {\em Proceedings of the IEEE conference on computer vision and
  pattern recognition}, pages 770--778, 2016.

\bibitem{he2004multiscale}
Xuming He, Richard~S Zemel, and Miguel~{\'A} Carreira-Perpi{\~n}{\'a}n.
\newblock Multiscale conditional random fields for image labeling.
\newblock In {\em Proceedings of the 2004 IEEE Computer Society Conference on
  Computer Vision and Pattern Recognition, 2004. CVPR 2004.}, volume~2, pages
  II--II. IEEE, 2004.

\bibitem{univperturb}
Jan Hendrik~Metzen, Mummadi Chaithanya~Kumar, Thomas Brox, and Volker Fischer.
\newblock Universal adversarial perturbations against semantic image
  segmentation.
\newblock In {\em Proceedings of the IEEE International Conference on Computer
  Vision}, pages 2755--2764, 2017.

\bibitem{pix2pix}
Phillip Isola, Jun-Yan Zhu, Tinghui Zhou, and Alexei~A Efros.
\newblock Image-to-image translation with conditional adversarial networks.
\newblock In {\em Proceedings of the IEEE conference on computer vision and
  pattern recognition}, pages 1125--1134, 2017.

\bibitem{kingma2014adam}
Diederik~P Kingma and Jimmy Ba.
\newblock Adam: A method for stochastic optimization.
\newblock {\em arXiv preprint arXiv:1412.6980}, 2014.

\bibitem{kohli2009robust}
Pushmeet Kohli, Philip~HS Torr, et~al.
\newblock Robust higher order potentials for enforcing label consistency.
\newblock {\em International Journal of Computer Vision}, 82(3):302--324, 2009.

\bibitem{krahenbuhl2011efficient}
Philipp Kr{\"a}henb{\"u}hl and Vladlen Koltun.
\newblock Efficient inference in fully connected crfs with gaussian edge
  potentials.
\newblock In {\em Advances in neural information processing systems}, pages
  109--117, 2011.

\bibitem{kurakin2016adversarial}
Alexey Kurakin, Ian Goodfellow, and Samy Bengio.
\newblock Adversarial machine learning at scale.
\newblock {\em arXiv preprint arXiv:1611.01236}, 2016.

\bibitem{unifiednips2018}
Kimin Lee, Kibok Lee, Honglak Lee, and Jinwoo Shin.
\newblock A simple unified framework for detecting out-of-distribution samples
  and adversarial attacks.
\newblock In {\em Advances in Neural Information Processing Systems}, pages
  7167--7177, 2018.

\bibitem{lis2019}
Krzysztof Lis, Krishna Nakka, Mathieu Salzmann, and Pascal Fua.
\newblock Detecting the unexpected via image resynthesis.
\newblock {\em arXiv preprint arXiv:1904.07595}, 2019.

\bibitem{parsenet2015}
Wei Liu, Andrew Rabinovich, and Alexander~C Berg.
\newblock Parsenet: Looking wider to see better.
\newblock {\em arXiv preprint arXiv:1506.04579}, 2015.

\bibitem{liu2016delving}
Yanpei Liu, Xinyun Chen, Chang Liu, and Dawn Song.
\newblock Delving into transferable adversarial examples and black-box attacks.
\newblock {\em arXiv preprint arXiv:1611.02770}, 2016.

\bibitem{fcn2015}
Jonathan Long, Evan Shelhamer, and Trevor Darrell.
\newblock Fully convolutional networks for semantic segmentation.
\newblock In {\em Proceedings of the IEEE conference on computer vision and
  pattern recognition}, pages 3431--3440, 2015.

\bibitem{lid2018}
Xingjun Ma, Bo Li, Yisen Wang, Sarah~M Erfani, Sudanthi Wijewickrema, Grant
  Schoenebeck, Dawn Song, Michael~E Houle, and James Bailey.
\newblock Characterizing adversarial subspaces using local intrinsic
  dimensionality.
\newblock {\em arXiv preprint arXiv:1801.02613}, 2018.

\bibitem{moosavi2017universal}
Seyed-Mohsen Moosavi-Dezfooli, Alhussein Fawzi, Omar Fawzi, and Pascal
  Frossard.
\newblock Universal adversarial perturbations.
\newblock In {\em Proceedings of the IEEE conference on computer vision and
  pattern recognition}, pages 1765--1773, 2017.

\bibitem{moosavi2017analysis}
Seyed-Mohsen Moosavi-Dezfooli, Alhussein Fawzi, Omar Fawzi, Pascal Frossard,
  and Stefano Soatto.
\newblock Analysis of universal adversarial perturbations.
\newblock {\em arXiv preprint arXiv:1705.09554}, 2017.

\bibitem{moosavi2016deepfool}
Seyed-Mohsen Moosavi-Dezfooli, Alhussein Fawzi, and Pascal Frossard.
\newblock Deepfool: a simple and accurate method to fool deep neural networks.
\newblock In {\em Proceedings of the IEEE conference on computer vision and
  pattern recognition}, pages 2574--2582, 2016.

\bibitem{nguyen2015deep}
Anh Nguyen, Jason Yosinski, and Jeff Clune.
\newblock Deep neural networks are easily fooled: High confidence predictions
  for unrecognizable images.
\newblock In {\em Proceedings of the IEEE conference on computer vision and
  pattern recognition}, pages 427--436, 2015.

\bibitem{nie2010efficient}
Feiping Nie, Heng Huang, Xiao Cai, and Chris~H Ding.
\newblock Efficient and robust feature selection via joint l2, 1-norms
  minimization.
\newblock In {\em Advances in neural information processing systems}, pages
  1813--1821, 2010.

\bibitem{papernot2016transferability}
Nicolas Papernot, Patrick McDaniel, and Ian Goodfellow.
\newblock Transferability in machine learning: from phenomena to black-box
  attacks using adversarial samples.
\newblock {\em arXiv preprint arXiv:1605.07277}, 2016.

\bibitem{saliencyattack}
Nicolas Papernot, Patrick McDaniel, Somesh Jha, Matt Fredrikson, Z~Berkay
  Celik, and Ananthram Swami.
\newblock The limitations of deep learning in adversarial settings.
\newblock In {\em 2016 IEEE European Symposium on Security and Privacy
  (EuroS\&P)}, pages 372--387. IEEE, 2016.

\bibitem{paszke2017automatic}
Adam Paszke, Sam Gross, Soumith Chintala, Gregory Chanan, Edward Yang, Zachary
  DeVito, Zeming Lin, Alban Desmaison, Luca Antiga, and Adam Lerer.
\newblock Automatic differentiation in pytorch.
\newblock 2017.

\bibitem{poursaeed2018generative}
Omid Poursaeed, Isay Katsman, Bicheng Gao, and Serge Belongie.
\newblock Generative adversarial perturbations.
\newblock In {\em Proceedings of the IEEE Conference on Computer Vision and
  Pattern Recognition}, pages 4422--4431, 2018.

\bibitem{ranjan2019attacking}
Anurag Ranjan, Joel Janai, Andreas Geiger, and Michael~J Black.
\newblock Attacking optical flow.
\newblock In {\em Proceedings of the IEEE International Conference on Computer
  Vision}, pages 2404--2413, 2019.

\bibitem{unet2015}
Olaf Ronneberger, Philipp Fischer, and Thomas Brox.
\newblock U-net: Convolutional networks for biomedical image segmentation.
\newblock In {\em International Conference on Medical image computing and
  computer-assisted intervention}, pages 234--241. Springer, 2015.

\bibitem{tramer2017ensemble}
Florian Tram{\`e}r, Alexey Kurakin, Nicolas Papernot, Ian Goodfellow, Dan
  Boneh, and Patrick McDaniel.
\newblock Ensemble adversarial training: Attacks and defenses.
\newblock {\em arXiv preprint arXiv:1705.07204}, 2017.

\bibitem{wen2016learning}
Wei Wen, Chunpeng Wu, Yandan Wang, Yiran Chen, and Hai Li.
\newblock Learning structured sparsity in deep neural networks.
\newblock In {\em Advances in neural information processing systems}, pages
  2074--2082, 2016.

\bibitem{SC2018}
Chaowei Xiao, Ruizhi Deng, Bo Li, Fisher Yu, Mingyan Liu, and Dawn Song.
\newblock Characterizing adversarial examples based on spatial consistency
  information for semantic segmentation.
\newblock In {\em Proceedings of the European Conference on Computer Vision
  (ECCV)}, pages 217--234, 2018.

\bibitem{DAG2017}
Cihang Xie, Jianyu Wang, Zhishuai Zhang, Yuyin Zhou, Lingxi Xie, and Alan
  Yuille.
\newblock Adversarial examples for semantic segmentation and object detection.
\newblock In {\em Proceedings of the IEEE International Conference on Computer
  Vision}, pages 1369--1378, 2017.

\bibitem{drn2017}
Fisher Yu, Vladlen Koltun, and Thomas Funkhouser.
\newblock Dilated residual networks.
\newblock In {\em Proceedings of the IEEE conference on computer vision and
  pattern recognition}, pages 472--480, 2017.

\bibitem{yuan2006model}
Ming Yuan and Yi Lin.
\newblock Model selection and estimation in regression with grouped variables.
\newblock {\em Journal of the Royal Statistical Society: Series B (Statistical
  Methodology)}, 68(1):49--67, 2006.

\bibitem{encnet}
Hang Zhang, Kristin Dana, Jianping Shi, Zhongyue Zhang, Xiaogang Wang, Ambrish
  Tyagi, and Amit Agrawal.
\newblock Context encoding for semantic segmentation.
\newblock In {\em Proceedings of the IEEE Conference on Computer Vision and
  Pattern Recognition}, pages 7151--7160, 2018.

\bibitem{psp2017}
Hengshuang Zhao, Jianping Shi, Xiaojuan Qi, Xiaogang Wang, and Jiaya Jia.
\newblock Pyramid scene parsing network.
\newblock In {\em Proceedings of the IEEE conference on computer vision and
  pattern recognition}, pages 2881--2890, 2017.

\bibitem{psanet2018}
Hengshuang Zhao, Yi Zhang, Shu Liu, Jianping Shi, Chen Change~Loy, Dahua Lin,
  and Jiaya Jia.
\newblock Psanet: Point-wise spatial attention network for scene parsing.
\newblock In {\em Proceedings of the European Conference on Computer Vision
  (ECCV)}, pages 267--283, 2018.

\end{thebibliography}
}
\clearpage

\section{Implementation Details}

In this section, we provide detailed explanations about the experiments described in Section 4 of the main paper.

\subsection{Models}
All models for the experiments were implemented in PyTorch~\cite{paszke2017automatic}. For generating adversarial attack, we use the advertorch~\cite{ding2019advertorch} library. Since different networks may have different normalization values for mean and standard deviation, we model normalization as a first layer inside the network and pass an RGB image scaled to the range [0,1].\\

\noindent{\textbf{FCN.}} We use the publicly released model\footnote{\url{https://github.com/hszhao/semseg}\label{psalink}} from the authors of~\cite{psanet2018}, which is trained together with PSANet~\cite{psanet2018} with an additional auxiliary loss. We use the ResNet-50 version for our evaluations.  \\

\noindent{\textbf{PSPNet.}} We use trained model\footref{psalink} released by the authors of~\cite{psanet2018}. It contains the same ResNet-50 as backbone network. The pyramid  pooling module is a 4-level pyramid, which is concatenated to the final convolutional spatial map and later fed to a classification layer.  \\

\noindent{\textbf{PSANet.}}We experiment with the trained model\footref{psalink} provided by authors of~\cite{psanet2018} with ResNet-50 as backbone network.  The PSA layer contains two sub-branches, namely collect and distribute, that favor a bi-directional information flow from each position to all other positions in the spatial feature map.\\

\noindent{\textbf{DANet.}} We use the trained model\footnote{\url{https://github.com/junfu1115/DANet}\label{danet}} from the authors of DANet~\cite{danet2019}. DANet uses ResNet-101 as backbone network followed by a spatial and channel wise attention module.
We use DANet with a hierarchy of grids of different sizes (4,8,16) in the last layer of each ResNet block. \\

\noindent{\textbf{DRN.}} We use the trained model\footnote{\url{https://github.com/fyu/drn}} released by authors of~\cite{drn2017}. We choose ResNet-22 as backbone network with dilated version corresponding to type \emph{D}.\\

\noindent{\textbf{U-Net.}} Along with the above-mentioned models, we evaluate the robustness of the U-Net architecture to local attacks. Due to the non-availability of a trained PyTorch~\cite{paszke2017automatic} version of the U-Net model, we re-trained it ourselves, achieving $33.7\%$ mIoU on Cityscapes.\\

Along with above six discussed models on Cityscapes, we experiment on PASCAL VOC~\cite{pascaldataset} with trained models of FCN~\cite{fcn2015}\footref{psalink} and PSANet~\cite{psanet2018}\footref{psalink} provided by authors of ~\cite{psanet2018}.

\subsection{Datasets}

\noindent{\textbf{Cityscapes}}:
We use the validation set of the Cityscapes~\cite{cityscapesdataset} dataset consisting of 500 images from 19 classes.  We divide the pixels at every position in the image into one of two sets, based on the category attribute provided by the authors. The first set consists of pixels belonging to static classes with category attribute \emph{ road, sidewalk, building, wall, fence, pole, traffic light, traffic sign, vegetation, terrain, sky}. The second set corresponds to regions of dynamic classes \emph{ person, rider, car, truck, bus, train, motorcycle, bicycle}.

The Cityscapes dataset has on average of $8\%$ of the pixels corresponding to dynamic classes in each image. Since our study was targeted to mis-classify the dynamic objects, images with  dynamic instances that occupy small regions might not be meaningful since such regions lie in the  immediate receptive field of their surroundings. Therefore, we take a subset of images consisting of 150 images whose combined region of instances corresponding to vehicle classes (\emph{ car, truck, bus, train, motorcycle, bicycle}) is greater than $8\%$. We provide the statistics of the resulting dataset in Table~\ref{tab:city_stats}.

\begin{table}[t]
	\centering
	{\small
		\resizebox{0.5\columnwidth}{!}{%
			\begin{tabular}{|l|c|}
				\midrule
				\multirow{1}{*}{Dynamic class}  & Number of Images \\
				\midrule
				Person & 115    \\ 
				Rider & 66\\
				Car & 150\\
				Truck & 33\\ 
				Bus & 23\\
				Train & 7\\
				 Motorcycle & 24\\
				 Bicycle & 88\\
				\bottomrule
			\end{tabular}%
		}
	}
\vspace{0.3cm}
	\caption{{\bf Cityscapes sampled dataset} statistics of the 150 images whose combined instance area of \emph{vehicle} categories is more than $8\%$. } 
	\label{tab:city_stats}
	\vspace{-0.5cm}
\end{table}

While the original Cityscapes dataset was captured at 2048 $\times$ 1024 resolution, we resize the image to half resolution of 1024 $\times$ 512 as the original size is too large to fit into GPU memory. Furthermore, we crop the bottom region of the image corresponding to the ego-vehicle of height $62$ pixels and resize the image back to 1024 $\times$ 512 pixels. For fair comparison, all models use the same 1024 $\times$ 512 resolution as input to the network without any tiling.\\

\noindent{\textbf{PASCAL VOC}}: We use a subset of 250 images from the original validation set consisting of 1449 images. It contains 20 foreground classes and one background class. In all settings, we target the pixels corresponding to all 20 foreground classes by perturbing a subset of the background area. 
\comment{We provide the statics of PASCAL VOC in Table~\ref{}.}

\subsection{Attack Algorithms}
We solve the indirect attacks given in Sections 3.1 and 3.2 of the main paper using the efficient iterative projected gradient descent algorithm~\cite{pgd2018} with an $\ell_p$-norm perturbation budget $\|\mathbf{\bM \odot \delta}\|_{p}<\epsilon$, where $p\in\{2,\infty\}$, using a step size $\alpha$. In all our experiments, we set the maximum $\ell_p$-norm of perturbation $\epsilon$ as the product of the number of iterations  given by 100  times $\alpha$ for $\ell_{\infty}$ attacks. 
For  $\ell_{2}$ attacks, we set the maximum $\ell_2$ norm of perturbation $\epsilon$ to 100. 

Formally, given an input image $\bX$, the adversarial attack minimizes the objective function, $J_{t}(\bX,\bM,\bF,\delta,\emph{f},\mathbf{y}^{pred},\mathbf{y}^t)$ to find the optimal $\delta$. We solve for $\delta$ in an iterative manner as
\begin{small}
	\vspace{-0.2cm}
\begin{align}
	\mathbf{\delta}^{(0)} &= \mathbf{0}\\
	\mathbf{\delta}^{(n+1)} &= \text{Clip}_{\mathbf{\varepsilon}}^p\left\{\mathbf{\delta}^{(n)} - \alpha \nabla_{\bX}J_{t}(\bX,\bM,\bF,\delta,\emph{f},\mathbf{y}^{pred},\mathbf{y}^t)\right\},
\end{align}
\end{small}
where $\text{Clip}_{\mathbf{\varepsilon}}^p$ clips the perturbation within the $\ell_p$ ball of radius $\epsilon$. For  $\ell_{\infty}$-norm based attacks, the gradient update is given by 
\begin{small}
	\begin{align}
	\nabla_{\bX}^{} J = \sign (\nabla_{\bX} (J_{t}(\bX,\bM,\bF,\delta,\emph{f},\mathbf{y}^{pred},\mathbf{y}^t))),
\end{align}
\end{small}
where $\sign$ is the signum function.\\

For $\ell_{2}$-norm based attacks, the gradient update is given by 
\begin{small}
	\begin{align}
	r  &= \nabla_{\bX} (J_{t}(\bX,\bM,\bF,\delta,\emph{f},\mathbf{y}^{pred},\mathbf{y}^t)) \\
		\nabla_{\bX}^{} J  &=  \frac{r}{\|r\|_{2}}.
	\end{align}
\end{small}

We observe that the DAG attack~\cite{DAG2017} is similar to the PGD-$\ell_2$ attack. While DAG projects the gradient by $\frac{r}{\|r\|_{\infty}}$, PGD-$\ell_2$ projects the gradient by $ \frac{r}{\|r\|_{2}}$. We emphasize  that our formalism for local indirect attacks is general and could be applied to other adversary generation techniques~\cite{DAG2017,carlini2017towards}.

\subsection{ Attack Detection Algorithms}

\noindent{\textbf{State-of-the-art methods.}}  In this paper, we compare the spatial consistency~\cite{SC2018} method and image re-synthesis method~\cite{lis2019} for adversarial attack detection at image level. In~\cite{SC2018},  given an input image of $1024 \times 512$ pixels, we crop 50 sufficiently overlapping pairs of patches of size  $256 \times 256$ and compute the average mIoU of the overlapped patch regions as the confidence score for attack detection.  In~\cite{lis2019}, we use the pix2pix generator to re-synthesize the image from the label map and then compute the $\ell_2$ distance of the input image and the re synthesized one in HOG feature space.\\

\noindent{\textbf{Our method.}} We provide the implementation details of our attack detection based on the Mahalanobis distance. To this end, we compute the class-conditional mean $\mu_{c}^\ell$  at every layer  $\ell$ of the network within locations corresponding to class label $c$ of the ground truth.   Furthermore, we compute the  group variance $\mathbf{\Sigma}^\ell$ for every layer $\ell$ of the network using features extracted at layer $\ell$. Since the number of features extracted on the training set can be high, we propose to compute the mean and variance on averaged features within locations corresponding to each label. 

Formally, let $\mathbf{\bX}_{j}^\ell$ be the feature extracted at layer $\ell$ at position $j$ for image $\bX$. Let the size of the feature map  $\mathbf{\bX}^\ell$ be given as $W_{\ell}$ $\times$ $H_{\ell}$ $\times$ $K_{\ell}$ where  $W_{\ell}$, $H_{\ell}$, $K_{\ell}$ are the  width, height and number of channels for layer $\ell$. Let  $L^{c}\in\mathbb{R}^{W_{\ell}\times H_{\ell}}$ be the  label mask activated at positions where the label is $c$, i.e., $L_j^c$ = 1 if the $j$-th pixel location belongs to label $c$ and $L_j^c = 0$ otherwise.  

First, we compute the averaged feature corresponding to label $c$ given by  $\mathbf{\hat \bX}_{c}^\ell$ = $\sum_{j | L_j=1} \mathbf{\bX}_{j}^\ell$. We then learn $\mu_{c}^\ell$ and $\mathbf{\Sigma}^\ell$  using $\{\mathbf{\hat \bX}_{c}^\ell | \bX \in [\bX_0,..., \bX_{N} ] \}$ extracted from all  $N$ images in the training set. In the end, we obtain $\mu_{c}^\ell$ $\in\mathbb{R}^{K_{\ell}}$ and $\mathbf{\Sigma}^\ell\in\mathbb{R^{K_{\ell}\times K_{\ell}}}$ for a layer $\ell$ in the network which is used as confidence score of Eq.(7) of main paper.

We extract features at the end of every block in the ResNet backbone followed by a context layer and a classification layer.  By doing so, we obtain a feature vector for the logistic detector of size $L = 6$ for FCN; $L = 7$ for PSANet; $L=7$ for PSPNet; $L=5$ for DANet; $L=5$ for DRN. For evaluation purpose, we use $80\%$ of the data for training and the remaining $20\%$ for testing.

\subsection{Performance Metrics}
For evaluation, we use the following metrics to measure the effectiveness of our indirect local attack.\\
{\tiny }
\noindent{\textbf{Intersection over Union.}} We report the mIoU used in the domain of segmentation to evaluate the effectiveness of the attack. 
\comment{We report the mIoU separately at  two positions: 1) at positions that we aim to fool (\emph{f});  2) at the remaining positions where the label should be preserved \emph{(p)}.}
We report the mIoU at positions that we aim to fool (\emph{f}) since at the rest of positions, the label is retained almost $98\%$ of times.
For untargeted attacks, we report $\text{mIoU}_{\mathbf{u}}^{\bfs}$ 
\comment{and $\text{mIoU}_{\bu}^{\bp}$}  as the mIoU calculated between the normal image prediction and its counterpart adversarial  image prediction at fooling \comment{and preserved} positions. \comment{respectively.} In the case of targeted attacks, along with  $\text{mIoU}_{\bu}^{\bfs}$,  \comment{and $\text{mIoU}_{\bu}^{\bp}$,} we report $\text{mIoU}_{\bt}^{\bfs}$ \comment{and $\text{mIoU}_{\bt}^{\bp}$} as the mIoU calculated between the normal image prediction and targeted label map at fooling \comment{and preserved positions respectively.} positions.\\

\noindent{\textbf{Attack Success Rate.}} We report the attack success rate at the percentage of pixels mis-classified/preserved relative to the total  number of pixels in the fooling/preserved positions, respectively. We report the mASR separately at  two positions: 1) at positions that we aim to fool (\emph{f});  2) at the remaining positions where the label should be preserved \emph{(p)}. We report $\text{mASR}_{\mathbf{u}}^{\bfs}$ and $\text{mASR}_{\bu}^{\bp}$ as the success rate calculated between the normal prediction and its adversarial image prediction at the fooling and preserved positions, respectively, for untargeted attacks. Specifically to calculate $\text{mASR}_{\mathbf{u}}^{\bfs}$, we assume the attack as  being successful at a pixel if it mis-classifies it to any label other than the normal predicted label.
In the case of targeted attacks, we additionally report $\text{mASR}_{\bt}^{\bfs}$ \comment{and $\text{mASR}_{\bt}^{\bp}$} as the success rate calculated between the normal prediction and targeted label map at fooling positions.  \comment{and preserved positions, respectively.}\\

\noindent{\textbf{Perceptibility.}}  We take the $\ell_{\infty}$-norm and  $\ell_{2}$-norm of the perturbation image as the two perceptibility scores.\\

We average the above metrics over the entire test set. Since in almost all experiments the labels are retained almost $>98\%$ times at preserved positions, we omitted reporting  $\text{mASR}_{u}^{\bp}$\comment{, $\text{mIoU}_{*}^{\bp}$} in the main paper.  We  reported  only $\text{mASR}_{\bt}^{\bfs}$  and  $\text{mIoU}_{\mathbf{u}}^{\bfs}$ at the fooling positions in the main paper as these metrics values are the most diverse in various attack settings.\\

\noindent{\textbf{AUROC.}}  The area under the receiver operating characteristic curve (AUROC) is computed by
 plotting the true positive rate (TPR) against the false positive rate (FPR)  by varying a threshold. We compute the AUROC both at image level and pixel level and report them in all perturbation settings.

\subsection{Cityscapes Experiments}
Table~\ref{tbl:Linf_full_city} and ~\ref{tbl:L2_full_city} shows the performance of different networks by varying noise levels for $\ell_{\infty}$ and $\ell_2$ attacks. Table~\ref{tbl:bound_full_linf} and~\ref{tbl:bound_full_l2} shows the impact of indirect attacks by perturbing static ones that are at-least distance $d$ pixels from dynamic object class with $\ell_{\infty}$ and $\ell_2$ attacks. Further, table~\ref{tbl:adaptive_full_city} shows the complete performance statistics of different networks by tuning the sparsity levels in adaptive attack strategy. We then show the impact of universal single fixed size patch attacks in Table~\ref{tbl:univ_full_city} by varying the patch size which is placed at the center of the image.

Finally, we show the attack detection results with  four perturbation settings: Global image perturbations (Global) to fool the entire image;  Universal patch perturbations (UP) at a fixed location to fool the entire image; Full static (FS) class perturbations to fool the dynamic classes; Adaptive patch (AP) perturbations in the static class regions to fool the dynamic objects. Table~\ref{tbl:attack_detection_full_fcn},~\ref{tbl:attack_detection_full_psp},~\ref{tbl:attack_detection_full_psa} and ~\ref{tbl:attack_detection_full_danet} shows the attack detection of our method and other two state-of-the-art detection methods discussed in main paper with FCN, PSP, PSANet, DANet respectively.

\subsubsection{Qualitative Results on Cityscapes}

Figure~\ref{fig:sup_indirect_city} visualizes adversarial images obtained by varying the step size $\alpha$ in both $\ell_{\infty}$ and $\ell_{2}$ for indirect local attacks with  PSANet~\cite{psanet2018}.
Figure~\ref{fig:sup_boundary_city} shows the outputs of indirect local attacks  by perturbing static class pixels that are at least a distance $d$ from a dynamic class pixel.
Figure~\ref{fig:sup_univ_city} shows the outputs of universal  patch attacks on different networks by varying the patch area in $\{1\%, 2.3\%, 4\%, 9\%\}$ of the image area. 
Figure~\ref{fig:sup_adaptive_city}  shows the results of adaptive local attacks on different networks by varying the sparsity level of the perturbation.\\

To understand the effectiveness of the Mahalanobis distance for attack detection, we visualize the internal subspaces of normal and adversarial samples. Figure~\ref{fig:viz_psp_city} and ~\ref{fig:viz_psa_city} show the visualizations of the nearest cluster assignment for each spatial location in the top-4 layers for PSPNet ~\cite{psp2017} and PSANet~\cite{psanet2018}, respectively. Figure~\ref{fig:detect_city1} the  output of pixel-level adversarial attack detection using mahalanobis distance  on PSANet~\cite{psanet2018} with adaptive indirect local attack at sparsity level $75\%$.


\begin{table*}[t]
\begin{center}
		{ 
			\resizebox{0.7\linewidth}{!}{%
				\begin{tabular}{@{}cc@{\hskip 0.3in}cc@{\hskip 0.3in}ccc@{\hskip 0.3in}cc@{}}
					\midrule
					\multirow{2}{*}{Networks}   
					& 	\multirow{2}{*}{ $\alpha$ }    
					&  \multicolumn{2}{c}{\centering mIoU}  
					& \multicolumn{3}{c}{\centering mASR} 
					& \multicolumn{2}{c}{Norm of $\delta$}  
					\\  \cmidrule(l{-5pt}r{16pt}){3-4} \cmidrule(l{-5pt}r{15pt}){5-7}  \cmidrule(l{-2pt}r){8-9}
 					& 
 					& $\text{mIoU}_{\mathbf{u}}^{\bfs}$  
 					& $\text{mIoU}_{\mathbf{t}}^{\bfs}$  
 					& $\text{mASR}_{\mathbf{u}}^{\bp}$ 
 					& $\text{mASR}_{\mathbf{u}}^{\bfs}$ 
 					& $\text{mASR}_{\mathbf{t}}^{\bfs}$ 
 					& $\ell_{\infty}$-norm 
 					& $\ell_{2}$-norm \\
 					\midrule
 					
 					\multirow{4}{*}{\begin{tabular}[c]{@{}c@{}} FCN~\cite{fcn2015} \end{tabular}} 
					& 1e-5 &0.65&0.08&100\%&6\%&5\%&0.001&0.83\\
				& 1e-4 &0.29&0.27&100\%&35\%&29\%&0.01&4.70 \\
				& 1e-3 &0.14&0.49&100\%&63\%&56\%&0.10&15.12\\
				& 5e-3 &0.11&0.55&100\%&69\%&62\%&0.40&50.93 \\
 					\midrule
					\multirow{4}{*}{\begin{tabular}[c]{@{}c@{}} PSPNet~\cite{psp2017} \end{tabular}} 
					& 1e-5 &0.71&0.10&99\%&15\%&12\%&0.001&0.77\\
				& 1e-4 & 0.06&0.53&100\%&98\%&86\%&0.01&3.10 \\
				& 1e-3 & 0.00& 0.62& 100\%&100\%&90\%&0.05&8.30\\
				& 5e-3 & 0.00&0.63&99\%&100\%&90\%&0.20 &37.99 \\
					\midrule
					\multirow{4}{*}{\begin{tabular}[c]{@{}c@{}} PSANet~\cite{psanet2018} \end{tabular}} 
					& 1e-5 &0.60&0.10&98\%&22\%&14\%&0.001&0.72\\
					& 1e-4 &0.04&0.51&99\%&99\%&86\%&0.01&2.68 \\
					& 1e-3 &0.01&0.60&99\%&100\%&90\%&0.05&8.10\\
					& 5e-3 &0.00&0.60&99\%&100\%&90\%&0.18&35.71 \\
					\midrule	
					\multirow{4}{*}{\begin{tabular}[c]{@{}c@{}} DANet~\cite{danet2019} \end{tabular}} 
					& 1e-5 &0.80&0.06&100\%&6\%&5\%&0.001&0.81\\
					& 1e-4 &0.11&0.50&99\%&91\%&80\%&0.01&3.90 \\
					& 1e-3 &0.01&0.65&99\%&99\%&90\%&0.04&8.30\\
					& 5e-3 &0.00&0.66&99\%&100\%&90\%&0.15&31.71 \\
					\midrule	\multirow{4}{*}{\begin{tabular}[c]{@{}c@{}} DRNet~\cite{drn2017} \end{tabular}} 
					& 1e-5 &0.64&0.09&99\%&9\%&6\%&0.001&0.87\\
					& 1e-4 &0.15&0.44&99\%&67\%&56\%&0.01&4.95 \\
					& 1e-3 &0.03&0.67&99\%&92\%&84\%&0.08&12.78\\
					& 5e-3 &0.02&0.67&99\%&94\%&87\%&0.27&40.2 \\
					\midrule
					\multirow{4}{*}{\begin{tabular}[c]{@{}c@{}} U-Net~\cite{unet2015} \end{tabular}} 
					& 1e-5 &0.35&0.15&99\%&29\%&20\%&0.001&0.91\\
					& 1e-4 &0.02&0.37&99\%&95\%&76\%&0.01&5.74 \\
					& 1e-3 &0.00&0.48&99\%&99\%&87\%&0.08&13.34\\
					& 5e-3 &0.00&0.52&99\%&100\%&89\%&0.28&38.89 \\				
 	 				\bottomrule
				\end{tabular}
			}
		}
	\end{center}
	\caption{ {\small \textbf{Indirect Attacks} on Cityscapes to fool dynamic classes while perturbing entire static ones with \emph{$\ell_{\infty}$} strategy}. The success rate of attacks increases with higher step size $\alpha$ although with higher perceptibility values. FCN is more robust to indirect attacks while PSANet and PSPNet are relatively more vulnerable to attack even at small step size like $\alpha$ = 1e-4.}
	\label{tbl:Linf_full_city}
\end{table*}

\begin{table*}[t]
	\begin{center}
	{ 
		\resizebox{0.7\linewidth}{!}{%
			\begin{tabular}{@{}cc@{\hskip 0.3in}cc@{\hskip 0.3in}ccc@{\hskip 0.3in}cc@{}}
				\midrule
			\multirow{2}{*}{Networks}   
			& 	\multirow{2}{*}{ $\alpha$ }    
			&  \multicolumn{2}{c}{\centering mIoU}  
			& \multicolumn{3}{c}{\centering mASR} 
			& \multicolumn{2}{c}{Norm of $\delta$}  
			\\  \cmidrule(l{-5pt}r{16pt}){3-4} \cmidrule(l{-5pt}r{15pt}){5-7}  \cmidrule(l{-2pt}r){8-9}
			& 
			& $\text{mIoU}_{\mathbf{u}}^{\bfs}$  
			& $\text{mIoU}_{\mathbf{t}}^{\bfs}$  
			& $\text{mASR}_{\mathbf{u}}^{\bp}$ 
			& $\text{mASR}_{\mathbf{u}}^{\bfs}$ 
			& $\text{mASR}_{\mathbf{t}}^{\bfs}$ 
			& $\ell_{\infty}$-norm 
			& $\ell_{2}$-norm \\
			\midrule
				\multirow{3}{*}{\begin{tabular}[c]{@{}c@{}} FCN~\cite{fcn2015} \end{tabular}} 
				& 8e-3 &0.60&0.10&100\%&13\%&10\%&0.02&0.58 \\
				& 4e-2 &0.36&0.21&99\%&33\%&26\%&0.05&1.75\\
				& 8e-2 & 0.27& 0.28&99\%& 44\%& 36\%&0.08&2.58 \\
				\midrule
				\multirow{3}{*}{\begin{tabular}[c]{@{}c@{}} PSPNet~\cite{psp2017} \end{tabular}} 
				& 8e-3 &0.68&0.12&99\%&24\%&20\%&0.01&0.51 \\
				& 4e-2 &0.23&0.37&99\%&81\%&67\%&0.02&1.28\\
				& 8e-2 & 0.02& 0.84&99\%& 96\%&91\%&0.03&1.17 \\
				\midrule
				\multirow{3}{*}{\begin{tabular}[c]{@{}c@{}} PSANet~\cite{psanet2018} \end{tabular}} 
				& 8e-3 &0.60&0.10&98\%&25\%&14\%&0.01&0.39 \\
				& 4e-2 &0.21&0.32&99\%&85\%&63\%&0.02&0.90\\
				& 8e-2 & 0.06& 0.53&99\%& 96\%&83\%&0.03&1.44 \\
				\midrule	\multirow{3}{*}{\begin{tabular}[c]{@{}c@{}} DANet~\cite{danet2019} \end{tabular}} 
			& 8e-3 &0.79&0.08&99\%&16\%&12\%&0.02&0.56 \\
			& 4e-2 &0.43&0.28&99\%&62\%&50\%&0.03&1.32\\
			& 8e-2 & 0.13& 0.54&99\%& 90\%&79\%&0.035&1.95 \\
				\midrule	\multirow{3}{*}{\begin{tabular}[c]{@{}c@{}} DRNet~\cite{drn2017} \end{tabular}} 
				& 8e-3 &0.63&0.10&99\%&16\%&10\%&0.02&0.65 \\
			& 4e-2 &0.24&0.37&99\%&60\%&48\%&0.06&2.14\\
			& 8e-2 & 0.13& 0.45&99\%& 76\%&65\%&0.08&3.02 \\
				\midrule
				\multirow{3}{*}{\begin{tabular}[c]{@{}c@{}} U-Net~\cite{unet2015} \end{tabular}} 
				& 8e-3 &0.32&0.17&99\%&36\%&25\%&0.02&0.70 \\
			& 4e-2 &0.05&0.32&98\%&85\%&66\%&0.08&2.76\\
			& 8e-2 & 0.02& 0.43&98\%& 95\%&79\%&0.09&3.43 \\
				\bottomrule
			\end{tabular}
		}
	}
\end{center}
	\caption{ {\small \textbf{Indirect Attacks} on Cityscapes to fool dynamic classes while perturbing entire static ones with  \emph{$ {\mathbf{\mathbf{\ell_{2}}}}$ strategy}. The perceptibility values of $\ell_2$ attack are much lower when compared to $\ell_{\infty}$ attack at a given success  rate of attack. Same as the case in $\ell_{\infty}$ attack, FCN is more robust to indirect attacks while PSANet and PSPNet are relatively more vulnerable to $\ell_2$ attacks.}}
	\label{tbl:L2_full_city}
	\vspace*{-0.6cm}
\end{table*}


\begin{table*}[t]
	\begin{center}
	{ 
	\resizebox{0.7\linewidth}{!}{%
		\begin{tabular}{@{}cc@{\hskip 0.3in}cc@{\hskip 0.3in}ccc@{\hskip 0.3in}cc@{}}
			\midrule
			\multirow{2}{*}{Networks}   
			& 	\multirow{2}{*}{ $d$ }    
			&  \multicolumn{2}{c}{\centering mIoU}  
			& \multicolumn{3}{c}{\centering mASR} 
			& \multicolumn{2}{c}{Norm of $\delta$}  
			\\  \cmidrule(l{-5pt}r{16pt}){3-4} \cmidrule(l{-5pt}r{15pt}){5-7}  \cmidrule(l{-2pt}r){8-9}
			& 
			& $\text{mIoU}_{\mathbf{u}}^{\bfs}$  
			& $\text{mIoU}_{\mathbf{t}}^{\bfs}$  
			& $\text{mASR}_{\mathbf{u}}^{\bp}$ 
			& $\text{mASR}_{\mathbf{u}}^{\bfs}$ 
			& $\text{mASR}_{\mathbf{t}}^{\bfs}$ 
			& $\ell_{\infty}$-norm 
			& $\ell_{2}$-norm \\
			\midrule
			
			\multirow{4}{*}{\begin{tabular}[c]{@{}c@{}} FCN~\cite{fcn2015} \end{tabular}} 
			& 50 &0.77&0.05&100\%&4\%&3\%&0.38&43.37\\
			& 100 &0.98&0.00&100\%&0\%&0\%&0.38&33.46\\
			& 150 &1.00&0.00&100\%&0\%&0\%&0.38&22.23\\
			\midrule
			\multirow{4}{*}{\begin{tabular}[c]{@{}c@{}} PSPNet~\cite{psp2017} \end{tabular}} 
		& 50 &0.14&0.37&99\%&96\%&74\%&0.28&41.83\\
		& 100 &0.24&0.26&98\%&86\%&60\%&0.29&33.00\\
		& 150 &0.55&0.12&97\%&35\%&23\%&0.34&22.86\\
			\midrule
			\multirow{4}{*}{\begin{tabular}[c]{@{}c@{}} PSANet~\cite{psanet2018} \end{tabular}} 
		& 50 &0.11&0.33&98\%&98\%&72\%&0.25&42.11\\
		& 100 &0.13&0.27&98\%&97\%&65\%&0.25&33.00\\
		& 150 &0.28&0.21&98\%&75\%&47\%&0.30&22.47\\
			\midrule	
			\multirow{4}{*}{\begin{tabular}[c]{@{}c@{}} DANet~\cite{danet2019} \end{tabular}} 
		& 50 &0.14&0.50&99\%&92\%&81\%&0.29&41.17\\
		& 100 &0.48&0.24&98\%&53\%&43\%&0.33&34.50\\
		& 150 &0.80&0.07&98\%&14\%&10\%&0.35&23.45\\
			\midrule	\multirow{4}{*}{\begin{tabular}[c]{@{}c@{}} DRNet~\cite{drn2017} \end{tabular}} 
			& 50 &0.37&0.20&99\%&34\%&22\%&0.43&46.30\\
			& 100 &0.73&0.05&99\%&5\%&3\%&0.44&37.24\\
			& 150 &0.94&0.00&100\%&0\%&0\%&0.47&25.87\\
			\midrule
			\multirow{4}{*}{\begin{tabular}[c]{@{}c@{}} U-Net~\cite{unet2015} \end{tabular}} 
		& 50 &0.01&0.25&98\%&97\%&70\%&0.43&44.62\\
		& 100 &0.03&0.20&96\%&90\%&60\%&0.47\%&39.61\\
		& 150 &0.10&0.17&95\%&74\%&47\%&0.49\%&33.27\\		
			\bottomrule
		\end{tabular}
	}
}
\end{center}
	\vspace*{0.2cm}
	\caption{ {\small \textbf{Impact of Local Attacks} by perturbing pixels that are at least $d$ pixels away from any dynamic class with \emph{$\ell_{\infty}$ strategy}. We observe PSANet~\cite{psanet2018} and UNet~\cite{unet2015} more vulnerable to indirect attacks when perturbations at large distances such $d=150$ while FCN~\cite{fcn2015} is barely effected.
}}
	\label{tbl:bound_full_linf}
\end{table*}

\begin{table*}[t]
	\begin{center}
	{ 
	\resizebox{0.7\linewidth}{!}{%
		\begin{tabular}{@{}cc@{\hskip 0.3in}cc@{\hskip 0.3in}ccc@{\hskip 0.3in}cc@{}}
			\midrule
			\multirow{2}{*}{Networks}   
			& 	\multirow{2}{*}{ $d$ }    
			&  \multicolumn{2}{c}{\centering mIoU}  
			& \multicolumn{3}{c}{\centering mASR} 
			& \multicolumn{2}{c}{Norm of $\delta$}  
			\\  \cmidrule(l{-5pt}r{16pt}){3-4} \cmidrule(l{-5pt}r{15pt}){5-7}  \cmidrule(l{-2pt}r){8-9}
			& 
			& $\text{mIoU}_{\mathbf{u}}^{\bfs}$  
			& $\text{mIoU}_{\mathbf{t}}^{\bfs}$  
			& $\text{mASR}_{\mathbf{u}}^{\bp}$ 
			& $\text{mASR}_{\mathbf{u}}^{\bfs}$ 
			& $\text{mASR}_{\mathbf{t}}^{\bfs}$ 
			& $\ell_{\infty}$-norm 
			& $\ell_{2}$-norm \\
			\midrule
			\multirow{4}{*}{\begin{tabular}[c]{@{}c@{}} FCN~\cite{fcn2015} \end{tabular}} 
		& 50 &0.80&0.05&100\%&3\%&3\%&0.31&10.71\\
		& 100 &0.98&0.00&100\%&0\%&0\%&0.32&9.95\\
		& 150 &1.00&0.00&100\%&0\%&0\%&0.40&9.43\\
		\midrule
		\multirow{4}{*}{\begin{tabular}[c]{@{}c@{}} PSPNet~\cite{psp2017} \end{tabular}} 
		& 50 &0.18&0.35&99\%&94\%&73\%&0.13&9.58\\
		& 100 &0.30&0.24&98\%&78\%&56\%&0.16&9.70\\
		& 150 &0.59&0.11&98\%&29\%&20\%&0.24&9.65\\
		\midrule
		\multirow{4}{*}{\begin{tabular}[c]{@{}c@{}} PSANet~\cite{psanet2018} \end{tabular}} 
		& 50 &0.10&0.37&99\%&98\%&76\%&0.19&9.41\\
		& 100 &0.14&0.29&98\%&95\%&67\%&0.22&9.43\\
		& 150 &0.31&0.21&98\%&70\%&45\%&0.27&9.55\\
		\midrule	
		\multirow{4}{*}{\begin{tabular}[c]{@{}c@{}} DANet~\cite{danet2019} \end{tabular}} 
		& 50 &0.27&0.40&99\%&83\%&72\%&0.19&9.90\\
		& 100 &0.67&0.15&98\%&33\%&26\%&0.22&9.87\\
		& 150 &0.85&0.05&98\%&10\%&7\%&0.30&9.51\\
		\midrule	\multirow{4}{*}{\begin{tabular}[c]{@{}c@{}} DRNet~\cite{drn2017} \end{tabular}} 
		& 50 &0.44&0.15&99\%&30\%&17\%&0.31&12.55\\
		& 100 &0.77&0.04&99\%&5\%&3\%&0.32&12.23\\
		& 150 &0.95&0.00&100\%&0\%&0\%&0.37&11.50\\
		\midrule
		\multirow{4}{*}{\begin{tabular}[c]{@{}c@{}} U-Net~\cite{unet2015} \end{tabular}} 
		& 50 &0.02&0.23&98\%&95\%&67\%&0.28&16.13\\
		& 100 &0.12&0.16&95\%&68\%&42\%&0.58&19.51\\
		& 150 &0.12&0.16&95\%&67\%&42\%&0.58&19.56\\		
			\bottomrule
		\end{tabular}
	}
}
\vspace{0.3cm}
\caption{ {\small \textbf{Impact of Local Attacks} by perturbing pixels that are at least d pixels away from any dynamic class with \emph{$\mathbf{\ell_{2}}$ strategy}. We observe PSANet~\cite{psanet2018} and UNet~\cite{unet2015} more vulnerable to indirect attacks when perturbations at large distances such $d=150$ while FCN~\cite{fcn2015} is barely effected.
}}
\label{tbl:bound_full_l2}
\end{center}
\end{table*}


\begin{table*}[t]
	\begin{center}
		{ 
			\resizebox{0.6\linewidth}{!}{%
					\begin{tabular}{@{}cc@{\hskip 0.3in}cc@{\hskip 0.3in}ccc@{\hskip 0.3in}cc@{}}
					\midrule
					\multirow{2}{*}{Networks}   
					& 	\multirow{2}{*}{ Sparsity }    
					&  \multicolumn{2}{c}{\centering mIoU}  
					& \multicolumn{3}{c}{\centering mASR} 
					& \multicolumn{2}{c}{Norm of $\delta$}  
					\\  \cmidrule(l{-5pt}r{16pt}){3-4} \cmidrule(l{-5pt}r{15pt}){5-7}  \cmidrule(l{-2pt}r){8-9}
					& 
					& $\text{mIoU}_{\mathbf{u}}^{\bfs}$  
					& $\text{mIoU}_{\mathbf{t}}^{\bfs}$  
					& $\text{mASR}_{\mathbf{u}}^{\bp}$ 
					& $\text{mASR}_{\mathbf{u}}^{\bfs}$ 
					& $\text{mASR}_{\mathbf{t}}^{\bfs}$ 
					& $\ell_{\infty}$-norm 
					& $\ell_{2}$-norm \\
					\midrule
 					
 					\multirow{4}{*}{\begin{tabular}[c]{@{}c@{}} FCN~\cite{fcn2015} \end{tabular}} 
 				& $75\%$ &0.52&0.12&100\%&18\%&13\%&0.15 & 4.04\\
 				& $85\%$ &0.67&0.07&100\%&9\%&6\%&0.14 &3.11\\
 				& $90\%$ &0.73&0.05&100\%&6\%&4\%&0.12& 2.54\\
 				& $95\%$ &0.84&0.03&100\%&2\%&2\%& 0.10 & 1.78\\
 				\midrule
				\multirow{4}{*}{\begin{tabular}[c]{@{}c@{}} PSPNet~\cite{psp2017} \end{tabular}} 
				& $75\%$ &0.19&0.38&99\%&89\%&71\%&0.09 & 4.87\\
			& $85\%$ &0.32&0.28&98\%&74\%&55\%&0.11&5.25 \\
			& $90\%$ &0.42&0.21&98\%&60\%&42\%&0.13 &5.30\\
			& $95\%$ &0.60&0.11&98\%&33\%&22\%&0.15 & 4.85\\
				\midrule
					\multirow{4}{*}{\begin{tabular}[c]{@{}c@{}} PSANet~\cite{psanet2018} \end{tabular}} 
				& $75\%$ &0.10&0.44&99\%&97\%&79\%&0.09&4.76 \\
			& $85\%$ &0.16&0.38&98\%&94\%&71\%&0.10& 5.20 \\
			& $90\%$ &0.20&0.32&98\%&89\%&64\%&0.12 & 5.19\\
			& $95\%$ &0.36&0.22&98\%&70\%&44\%&0.14 &5.07\\
				\midrule	\multirow{4}{*}{\begin{tabular}[c]{@{}c@{}} DANet~\cite{danet2019} \end{tabular}} 
				& $75\%$ &0.30&0.37&99\%&78\%&65\%&0.12 &5.63\\
			& $85\%$ &0.49&0.23&99\%&57\%&46\%&0.14 &5.79\\
			& $90\%$ &0.64&0.16&99\%&40\%&30\%&0.15&5.80\\
			& $95\%$ &0.71&0.12&99\%&29\%&21\%&0.13 &3.95\\
				\midrule	\multirow{4}{*}{\begin{tabular}[c]{@{}c@{}} DRNet~\cite{drn2017} \end{tabular}} 
					& $75\%$ &0.42&0.19&100\%&35\%&22\%&0.18 & 5.40\\
				& $85\%$ &0.55&0.11&100\%&22\%&13\%&0.15 &4.43\\
				& $90\%$ &0.63&0.08&100\%&15\%&10\%&0.14&3.84\\
				& $95\%$ &0.77&0.05&100\%&8\%&5\%&0.13 &2.81\\
				\midrule
				\multirow{4}{*}{\begin{tabular}[c]{@{}c@{}} U-Net~\cite{unet2015} \end{tabular}} 
				& $75\%$ &0.12&0.20&96\%&70\%&44\%&0.15 &6.56\\
				& $85\%$ &0.19&0.15&96\%&52\%&32\%&0.19& 6.81 \\
				& $90\%$ &0.25&0.13&96\%&42\%&25\%&0.22&6.54\\
				& $95\%$  &0.36&0.11&96\%&27\%&16\%&0.23 &5.73\\
 	 					\bottomrule
				\end{tabular}
			}
		}
\end{center}
	\caption{ {\small \textbf{Adaptive Indirect Local Attacks} on Cityscapes. We compute the performance statistics for different sparsity levels of perturbation region. By enforcing group sparsity prior, we can attack context-aware networks such as PSANet~\cite{psanet2018}, PSPNet~\cite{psp2017} and DANet~\cite{danet2019} at relatively higher success rate than baseline FCN~\cite{fcn2015}.}}
	\label{tbl:adaptive_full_city}
\end{table*}

\begin{table*}[t]
	\begin{center}
	{ 
		\resizebox{0.5\linewidth}{!}{%
			\begin{tabular}{@{}cc@{\hskip 0.5in}c@{\hskip 0.3in}c@{\hskip 0.3in}cc@{}}
				\midrule
				\multirow{2}{*}{Networks}   
				& 	\multirow{2}{*}{\begin{tabular}[c]{@{}c@{}} Patch size \\{$h$$\times$$w$  (\text{area}\%)}   \end{tabular}}
				&  \multicolumn{1}{l}{mIoU}  
				& \multicolumn{1}{l}{mASR} 
				& \multicolumn{2}{c}{Norm of $\delta$}  
				\\  \cmidrule(l{-5pt}r{16pt}){3-3} \cmidrule(l{-5pt}r{15pt}){4-4}  \cmidrule(l{-2pt}r){5-6}
				& 
				&   $\text{mIoU}_{\mathbf{u}}^{\bfs}$  
				& $\text{mASR}_{\mathbf{u}}^{\bfs}$ 

				& $\ell_{\infty}$-norm 
				& $\ell_{2}$-norm \\ 
				\midrule
				\multirow{4}{*}{\begin{tabular}[c]{@{}c@{}} FCN~\cite{fcn2015} \end{tabular}} 
				& $51 \times 102$  ${\tiny }\mathbf{ (1.0\%)}$ &0.86&2\%&0.30&25.36 \\
				& $76 \times 157$ $ \mathbf{ (2.3\%)}$ &0.78&4\%&0.30&37.60 \\
				& $102 \times 204$ $\mathbf{ (4.0\%)}$ &0.73&10\%&0.30&51.80\\
				& $153\times 306$  $\mathbf{ (9.0\%) }$  &0.58&18\%&0.30&78.32 \\
				\midrule
				\multirow{4}{*}{\begin{tabular}[c]{@{}c@{}} PSPNet~\cite{psp2017} \end{tabular}} 
				& $51 \times 102$  ${\tiny }\mathbf{ (1.0\%)}$ &0.80&3\%&0.30&25.52 \\
			& $76 \times 157$ $ \mathbf{ (2.3\%)}$ &0.63&10\%&0.30&38.43\\
			& $102 \times 204$ $\mathbf{ (4.0\%)}$ &0.44&27\%&0.30&50.32\\
			& $153\times 306$  $\mathbf{ (9.0\%) }$  &0.09&84\%&0.30&74.92 \\
				\midrule
				\multirow{4}{*}{\begin{tabular}[c]{@{}c@{}} PSANet~\cite{psanet2018} \end{tabular}} 
				& $51 \times 102$  ${\tiny }\mathbf{ (1.0\%)}$ &0.41&38\%&0.30&26.69 \\
			& $76 \times 157$ $ \mathbf{ (2.3\%)}$ &0.23&60\%&0.30&38.60 \\
			& $102 \times 204$ $\mathbf{ (4.0\%)}$ &0.14&71\%&0.30&50.39\\
			& $153\times 306$  $\mathbf{ (9.0\%) }$  &0.04&90\%&0.30&78.02 \\
				\midrule	\multirow{4}{*}{\begin{tabular}[c]{@{}c@{}} DANet~\cite{danet2019} \end{tabular}} 
				& $51 \times 102$  ${\tiny }\mathbf{ (1.0\%)}$ &0.79&4\%&0.30&26.45 \\
			& $76 \times 157$ $ \mathbf{ (2.3\%)}$ &0.71&10\%&0.30&37.24 \\
			& $102 \times 204$ $\mathbf{ (4.0\%)}$ &0.65&15\%&0.30&49.86\\
			& $153\times 306$  $\mathbf{ (9.0\%) }$  &0.40&42\%&0.30&74.60 \\
				\midrule	\multirow{4}{*}{\begin{tabular}[c]{@{}c@{}} DRNet~\cite{drn2017} \end{tabular}} 
				& $51 \times 102$  ${\tiny }\mathbf{ (1.0\%)}$ &0.82&2\%&0.30&26.28 \\
			& $76 \times 157$ $ \mathbf{ (2.3\%)}$ &0.77&7\%&0.30&39.27 \\
			& $102 \times 204$ $\mathbf{ (4.0\%)}$ &0.70&14\%&0.30&52.23\\
			& $153\times 306$  $\mathbf{ (9.0\%) }$  &0.55&28\%&0.30&78.32 \\
				\midrule
				\multirow{4}{*}{\begin{tabular}[c]{@{}c@{}} U-Net~\cite{unet2015} \end{tabular}} 
				& $51 \times 102$  ${\tiny }\mathbf{ (1.0\%)}$ &0.32&26\%&0.30&29.95 \\
			& $76 \times 157$ $ \mathbf{ (2.3\%)}$ &0.13&58\%&0.30&44.42 \\
			& $102 \times 204$ $\mathbf{ (4.0\%)}$ &0.06&76\%&0.30&58.15\\
			& $153\times 306$  $\mathbf{ (9.0\%) }$  &0.02&90\%&0.30&86.06 \\
				\bottomrule
			\end{tabular}
		}
	}
\end{center}
	\caption{ {\small \textbf{Universal Local Attacks} on Cityscapes by tuning the  patch size {$h$$\times$$w$  (\text{area}\%)}  on different networks. PSANet~\cite{psanet2018} and UNet~\cite{unet2015} are highly sensitive to patch attacks even when the patch is $~1\%$ of image area. Note that the attack is untargeted and aimed to fool entire scene by placing a fixed size at the center of image. We use $\ell_{\infty}$ based attack with $\alpha=0.001$ and $\epsilon=0.3$. }}
	\label{tbl:univ_full_city}
\end{table*}


\begin{table*}[t]
	\begin{center}
			\resizebox{0.9\linewidth}{!}{%
			\begin{tabular}{ccccccccc}
				\midrule
				\multirow{2}{*}{Networks}   
								& 	\multirow{2}{*}{\begin{tabular}[c]{@{}c@{}} Perturbation \\ region \end{tabular}}
				& 	\multirow{2}{*}{\begin{tabular}[c]{@{}c@{}} Fooling \\ region \end{tabular}}
				& {}
				&\multicolumn{2}{c}{Norm of $\delta$}  
				& 	\multirow{2}{*}{\begin{tabular}[c]{@{}c@{}} Misclassified \\pixels $\%$  \end{tabular}}
				& Global AUROC 
				& Local AUROC  \\  
				\cmidrule{5-6} \cmidrule(l{7pt}r{5pt}){8-8}\cmidrule(l{7pt}r){9-9}
				& 
				& 
				& 
				& $\ell_{\infty}$ 
				& $\ell_2$  
				& 
				& { SC~\cite{SC2018} / Re-Syn~\cite{lis2019} / \ours} 
				&\ours \\
				\midrule
				
				\multirow{24}{*}{FCN \cite{fcn2015}} &
				\multirow{1}{*}{Global} &
				\multirow{1}{*}{\full} & &
				 0.09& 17.67 & $91 \%$ & \textbf{1.00 / 1.00} / {0.94}& 0.90  \\
				\cmidrule{2-9}
				
					 &
		\multirow{7}{*}{\fs} &
		\multirow{7}{*}{\dyn} & & 0.001
		& 0.83 &1\%& 0.48 / 0.53 / \textbf{0.89}&  0.80 \\
		& && &0.01&4.70 &5\%&0.54 / 0.67 / {\bf 1.00}& 0.83  \\	
		&  &&&0.10&15.12&9\%& 0.65 / 0.75 / \textbf{1.00}& 0.83  \\	
		&&&&0.40&50.93  &10\%& 0.93 / 0.76 / \textbf{1.00}&  0.73 \\
		\cmidrule{4-9}
		 &&& & 0.02
		 & 0.58 &  2\% & {0.51 / 0.56} / \textbf{0.58}& 0.83  \\
		&  & & &0.05& 1.75 &  5\%& {0.55 / 0.67} / \textbf{0.82}& 0.86  \\
		&				 & &    & 0.08   & 2.58 &  6\% & {0.57/ 0.71} / \textbf{0.90}& 0.87  \\

		\cmidrule{2-9}
		
		 &
		\multirow{4}{*}{\up} &
		\multirow{4}{*}{\full} 
		& & 0.30 & 25.46 & 2\% & {0.70 / 0.55} / \textbf{0.88}& 0.96  \\
		&  & &&0.30 & 37.60   & $4 \%$ & {0.82 / 0.64} / \textbf{1.00}& 0.94  \\
		&				 & &        &0.30& 51.80 &  $10 \%$ & {0.90 / 0.75} / {1.00}& 0.94  \\
		&	 		&		  &       &0.30 & 7.32  & $18 \%$ & {0.99 / 0.94} / \textbf{1.00}& 0.95  \\
		\cmidrule{4-9}
		\cmidrule{2-9}

		&
		\multirow{4}{*}{\ap} &
		\multirow{4}{*}{\dyn} 
		& & 0.15 & 4.04 & $3 \%$ & {0.68 / 0.65} / \textbf{0.92}& 0.88  \\
		&  & & &  0.14& 3.11 & $2 \%$ & {0.61 / 0.57} / \textbf{0.87}& 0.89  \\
		&				 & &        &0.12  &2.54  & $1 \%$ & {0.60 / 0.55} / \textbf{0.80}& 0.90  \\
		&	 		&		  &       & 0.10& 1.78  & $1 \%$ & {0.60 / 0.52} / \textbf{0.73}& 0.91  \\
		\midrule		 
				 				 				 				 

			\end{tabular}
		}
\end{center}
\vspace*{0.1cm}
\caption{ {\small \textbf{Attack detection} on Cityscapes with different perturbation settings on \textbf{FCN~\cite{fcn2015}}. We perform mahalanobis based attack detection in four settings namely Global:Global image perturbations, UP:Universal patch perturbations; FS : Full static class perturbations. . We tune the noise level or patch size or sparsity levels of attack generation process to achieve different range of success rate. As observed, SC~\cite{SC2018} and Re-Syn~\cite{lis2019}   perform well only when large percentage of pixels are misclassified while we outperform them by a large margin in all other settings.}}
\label{tbl:attack_detection_full_fcn}
\end{table*}

\begin{table*}[t]
	\begin{center}
		\resizebox{0.9\linewidth}{!}{%
			\begin{tabular}{ccccccccc}
				\midrule
				\multirow{2}{*}{Networks}   
				& 	\multirow{2}{*}{\begin{tabular}[c]{@{}c@{}} Perturbation \\ region \end{tabular}}
				& 	\multirow{2}{*}{\begin{tabular}[c]{@{}c@{}} Fooling \\ region \end{tabular}}
				& {}
				&\multicolumn{2}{c}{Norm of $\delta$}  
				& 	\multirow{2}{*}{\begin{tabular}[c]{@{}c@{}} Misclassified \\pixels $\%$  \end{tabular}}
				& Global AUROC 
				& Local AUROC  \\  
				\cmidrule{5-6} \cmidrule(l{7pt}r{5pt}){8-8}\cmidrule(l{7pt}r){9-9}
				& 
				& 
				& 
				& $\ell_{\infty}$ 
				& $\ell_2$  
				& 
				& { SC~\cite{SC2018} / Re-Syn~\cite{lis2019} / \ours} 
				&\ours \\
				\midrule
				
				\multirow{24}{*}{PSPNet \cite{psp2017}} &
				\multirow{1}{*}{Global} &
				\multirow{1}{*}{\full} & &
				  0.06 & 10.74  & $83 \%$ & {0.90} / \textbf{1.00} / {0.99}& 0.81  \\
				\cmidrule{2-9}
				&
				\multirow{7}{*}{\fs} &
				\multirow{7}{*}{\dyn} & &
				0.001 & 0.77 & $3 \%$ & {0.49 / 0.56} / \textbf{1.00}& 0.84  \\
				&  & & & 0.01 & 3.10 & $14 \%$ & {0.48 / 0.76} / \textbf{1.00}& 0.90  \\
				&				 & &        & 0.05 & 8.30 & $14 \%$ & {0.52 / 0.77} / \textbf{1.00}& 0.85  \\
				&	 		&		  &       & 0.20 & 37.99 & $14 \%$ & {0.88 / 0.78} / \textbf {1.00}& 0.88  \\
				\cmidrule{4-9}
				&&& &
				0.01 & 0.51 & $4 \%$ & {0.50 / 0.59} / \textbf{1.00}& 0.85  \\
				&  & & & 0.02 & 1.28 & $12 \%$ & {0.52 / 0.72} / \textbf {1.00}& 0.87  \\
				&				 & &        & 0.03 & 1.17 & $14 \%$ & {0.52 / 0.73} / \textbf{1.00}& 0.87  \\

				\cmidrule{2-9}
				
				&
				\multirow{4}{*}{\up} &
				\multirow{4}{*}{\full} 
				& &
				0.30 & 25.52 & $3 \%$ & {0.57 / 0.55} / \textbf{1.00}& 0.93  \\
				&  & & & 0.30 & 38.43 & $10 \%$ & {0.62 / 0.70} / \textbf{1.00}& 0.96  \\
				&				 & &        & 0.30 & 50.32 & $27 \%$ & {0.65 / 0.89} / \textbf{1.00}& 0.96  \\
				&	 		&		  &       & 0.30 & 74.92 & $84 \%$ & {0.87 / 1.00} / {1.00}& 0.97  \\
				\cmidrule{4-9}				
				&
				\multirow{4}{*}{\ap} &
				\multirow{4}{*}{\dyn} & &
				0.09 & 4.87 & $12 \%$ & {0.65 / 0.82} / \textbf{0.99}& 0.90  \\
				&  & & & 0.11 & 5.25 & $10 \%$ & {0.59 / 0.76} / \textbf{0.98}& 0.82  \\
				&				 & &        & 0.13 & 5.30 & $9 \%$ & {0.56 / 0.72} / \textbf{0.99}& 0.82  \\
				&	 		&		  &       & 0.15 & 4.85 & 5 \% & {0.55 / 0.69} / \textbf{1.00}& 0.84  \\
				\midrule		 
				
				
			\end{tabular}
		}
	\end{center}
	\vspace*{0.1cm}
	\caption{ {\small \textbf{Attack detection} on Cityscapes with different perturbation settings on \textbf{PSPNet~\cite{psp2017}}. We perform mahalanobis based attack detection in four settings namely Global:Global image perturbations, UP:Universal patch perturbations; FS : Full static class perturbations. We tune the noise level or patch size or sparsity levels of attack generation process to achieve different range of success rate. As observed, SC~\cite{SC2018} and Re-Syn~\cite{lis2019}   perform well only when large percentage of pixels are misclassified while we outperform them by a large margin in all other settings.}}
	\label{tbl:attack_detection_full_psp}
\end{table*}

\begin{table*}[t]
	\begin{center}
		\resizebox{0.9\linewidth}{!}{%
			\begin{tabular}{ccccccccc}
				\midrule
				\multirow{2}{*}{Networks}   
				& 	\multirow{2}{*}{\begin{tabular}[c]{@{}c@{}} Perturbation \\ region \end{tabular}}
				& 	\multirow{2}{*}{\begin{tabular}[c]{@{}c@{}} Fooling \\ region \end{tabular}}
				& {}
				&\multicolumn{2}{c}{Norm of $\delta$}  
				& 	\multirow{2}{*}{\begin{tabular}[c]{@{}c@{}} Misclassified \\pixels $\%$  \end{tabular}}
				& Global AUROC 
				& Local AUROC  \\  
				\cmidrule{5-6} \cmidrule(l{7pt}r{5pt}){8-8}\cmidrule(l{7pt}r){9-9}
				& 
				& 
				& 
				& $\ell_{\infty}$ 
				& $\ell_2$  
				& 
				& { SC~\cite{SC2018} / Re-Syn~\cite{lis2019} / \ours} 
				&\ours \\
				\midrule
				\multirow{24}{*}{PSANet \cite{psanet2018}} &
				\multirow{1}{*}{Global} &
				\multirow{1}{*}{\full} & &
				0.04 & 8.26 & $93 \%$ & {0.90} / \textbf{1.00} / {0.94}& 0.75  \\
				\cmidrule{2-9}
				&
				\multirow{7}{*}{\fs} &
				\multirow{7}{*}{\dyn} & &
				0.001 & 0.72 & $4 \%$ & {0.49 / 0.56} / \textbf{1.00}& 0.88  \\
				&  & & & 0.01 & 2.68 & $14 \%$ & {0.48 / 0.77} / \textbf{1.00}& 0.92  \\
				&				 & &        & 0.05 & 8.10 & $14 \%$ & {0.50 / 0.78} / \textbf{1.00}& 0.89  \\
				&	 		&		  &       & 0.18 & 35.71 & $14 \%$ & {0.87 / 0.78} / \textbf{1.00}& 0.87  \\
				\cmidrule{4-9}
				&  & & & 0.01 & 0.39 & $4 \%$ & {0.51 / 0.57} / \textbf{1.00}& 0.88  \\
				&& &   & 0.02 & 0.90 & $13 \%$ & {0.49 / 0.73} / \textbf{1.00}& 0.92  \\
				&	&  &    & 0.03 & 1.44 & $14 \%$ & {0.49 / 0.77} / \textbf{1.00}& 0.92  \\
				\cmidrule{2-9}
				&
				\multirow{4}{*}{\up} &
				\multirow{4}{*}{\full} 
				& &
				0.30 & 26.69 & $38 \%$ & {0.60 / 1.00} / {1.00}& 0.99  \\
				&  & & & 0.30 & 38.60 & $60 \%$ & {0.62 / 1.00} / {1.00}& 0.98  \\
				&				 & &        & 0.30 & 50.39 & $71 \%$ & {0.69 / 1.00} / {1.00}& 0.97  \\
				&	 		&		  &       & 0.30 & 78.02 & $90 \%$ & {0.85 / 1.00} / {1.00}& 0.98  \\
				\cmidrule{4-9}
				&
				\multirow{4}{*}{\ap} &
				\multirow{4}{*}{\dyn} & &
				0.09 & 4.76 & $14 \%$ & {0.54 / 0.85} / \textbf{1.00}& 0.95  \\
				&  & & & 0.10 & 5.20 & $14 \%$ & {0.52 / 0.83} / \textbf{1.00}& 0.94  \\
				&				 & &        & 0.12 & 5.19 & $13 \%$ & {0.54 / 0.81} / \textbf{1.00}& 0.92  \\
				&	 		&		  &       & 0.14 & 5.07 & $10 \%$ & {0.52 / 0.78} / \textbf{0.94}& 0.91  \\
				\midrule		 
			\end{tabular}
		}
	\end{center}
	\vspace*{0.1cm}
	\caption{ {\small \textbf{Attack detection} on Cityscapes with different perturbation settings  on \textbf{PSANet~\cite{psanet2018}}. We perform mahalanobis based attack detection in four settings namely Global:Global image perturbations, UP:Universal patch perturbations; FS : Full static class perturbations. . We tune the noise level or patch size or sparsity levels of attack generation process to achieve different range of success rate. As observed, SC~\cite{SC2018} and Re-Syn~\cite{lis2019}   perform well only when large percentage of pixels are misclassified while we outperform them by a large margin in all other settings. }}
	\label{tbl:attack_detection_full_psa}
\end{table*}

\begin{table*}[t]
	\begin{center}
		\resizebox{0.9\linewidth}{!}{%
			\begin{tabular}{ccccccccc}
				\midrule
				\multirow{2}{*}{Networks}   
				& 	\multirow{2}{*}{\begin{tabular}[c]{@{}c@{}} Perturbation \\ region \end{tabular}}
				& 	\multirow{2}{*}{\begin{tabular}[c]{@{}c@{}} Fooling \\ region \end{tabular}}
				& {}
				&\multicolumn{2}{c}{Norm of $\delta$}  
				& 	\multirow{2}{*}{\begin{tabular}[c]{@{}c@{}} Misclassified \\pixels $\%$  \end{tabular}}
				& Global AUROC 
				& Local AUROC  \\  
				\cmidrule{5-6} \cmidrule(l{7pt}r{5pt}){8-8}\cmidrule(l{7pt}r){9-9}
				& 
				& 
				& 
				& $\ell_{\infty}$ 
				& $\ell_2$  
				& 
				& { SC~\cite{SC2018} / Re-Syn~\cite{lis2019} / \ours} 
				&\ours \\
				\midrule
				
				\multirow{24}{*}{DANet \cite{danet2019}} &
				\multirow{1}{*}{Global} &
				\multirow{1}{*}{\full} & &
				0.06 & 12.55 & $82 \%$ & {0.89} / \textbf{1.00} / \textbf{1.00}& 0.68  \\
				\cmidrule{2-9}
				&
				\multirow{7}{*}{\fs} &
				\multirow{7}{*}{\dyn} 
				& &0.01 & 0.81 & $1 \%$ & {0.50 / 0.51} / \textbf{0.64}& 0.88  \\
				&  & & & 0.01 & 3.90 & $14 \%$ & {0.52 / 0.72} / \textbf{0.96}& 0.86  \\
				&				 & &        & 0.04 & 8.30 & $14 \%$ & {0.56 / 0.74} / \textbf{0.99}& 0.92  \\
				&	 		&		  &       & 0.15 & 31.71 & $14 \%$ & {0.84 / 0.75} / \textbf{1.00}& 0.94  \\
				\cmidrule{4-9}
				&  & & & 0.02 & 0.56 & $3 \%$ & {0.50 / 0.54} / \textbf{0.67}& 0.86  \\
				&				 & &        & 0.03 & 1.32 & $9 \%$ & {0.48 / 0.64} / \textbf{0.89}& 0.86  \\
				&	 		&		  &       & 0.03 & 1.95 & $14 \%$ & {0.50 / 0.70} / \textbf{0.87}& 0.88  \\
				\cmidrule{2-9}
				&
				\multirow{4}{*}{\up} &
				\multirow{4}{*}{\full} 
				& &0.30 & 26.45 & $4 \%$ & {0.74 / 0.57} / \textbf{0.77}& 0.89  \\
				&  & & & 0.30 & 37.24 & $10 \%$ & {0.80 / 0.64} / \textbf{0.92}& 0.83  \\
				&				 & &        & 0.30 & 49.86 & $15 \%$ & {0.73 / 0.75} / \textbf{0.99}& 0.87  \\
				&	 		&		  &       & 0.30 & 74.60 & $42 \%$ & {0.88 / 0.92} / \textbf{1.00}& 0.89  \\
				\cmidrule{4-9}
				&
				\multirow{4}{*}{\ap} &
				\multirow{4}{*}{\dyn} 
				& &0.12 & 5.63 & $12 \%$ & {0.58 / 0.75} / \textbf{0.99}& 0.82  \\
				&  & & & 0.14 & 5.79 & $9 \%$ & {0.54 / 0.68} / \textbf{0.99}& 0.82  \\
				&				 & &        & 0.15 & 5.80 & $6 \%$ & {0.50 / 0.63} / \textbf{0.95}& 0.81  \\
				&	 		&		  &       & 0.13 & 3.95 & $5 \%$ & {0.51 / 0.58} / \textbf{0.85}& 0.83  \\
				\midrule		 
			\end{tabular}
		}
	\end{center}
	\vspace*{0.1cm}
	\caption{ {\small \textbf{Attack detection} on Cityscapes with different perturbation settings on \textbf{DANet~\cite{danet2019}}. We perform mahalanobis based attack detection in four settings namely Global:Global image perturbations, UP:Universal patch perturbations; FS : Full static class perturbations. . We tune the noise level or patch size or sparsity levels of attack generation process to achieve different range of success rate. As observed, SC~\cite{SC2018} and Re-Syn~\cite{lis2019}   perform well only when large percentage of pixels are misclassified while we outperform them by a large margin in all other settings.}}
	\label{tbl:attack_detection_full_danet}
\end{table*}


\providecommand{\localwidth}{}
\renewcommand{\localwidth}{0.5\linewidth}

\providecommand{\localheight}{}
\renewcommand{\localheight}{2cm}

\begin{figure*}[hbt!]
	\centering
	\includegraphics[width=\localwidth]{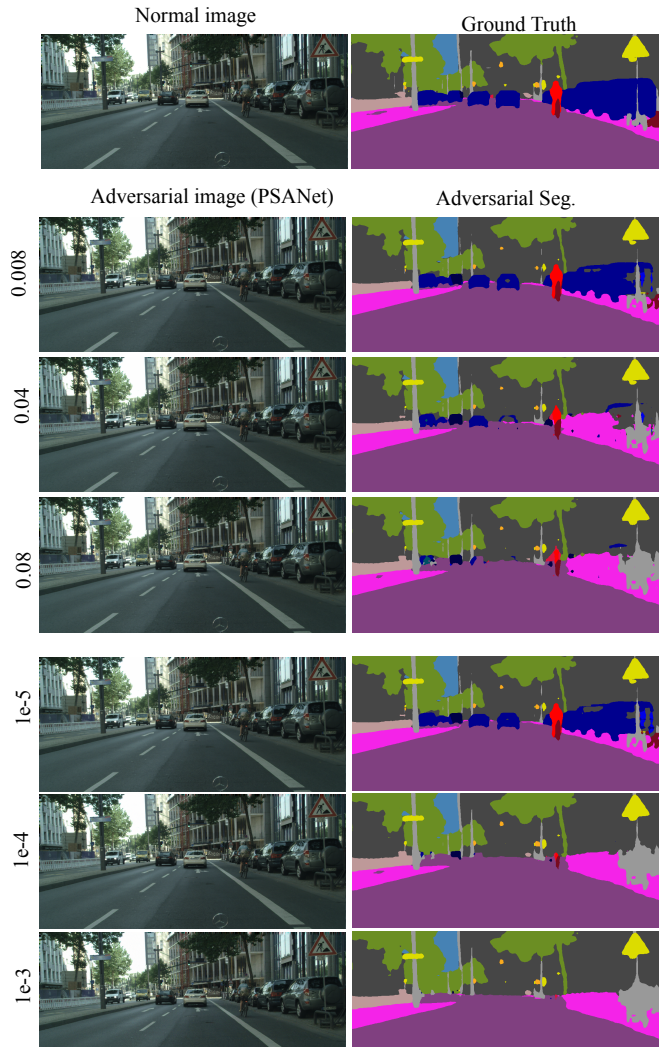}
		\vspace*{0.2cm}
	\caption{{\small {\bf Indirect Attacks} on Cityscapes  to fool dynamic classes while perturbing complete static ones using $\ell_2$ and $\ell_{\infty}$ attack. We use  $\alpha= \{8\text{e-}3, 4\text{e-}2,8\text{e-}2\}$ for $\ell_2$ attacks and $\alpha$=$\{1\text{e-}5, 1\text{e-}4,1\text{e-}3,5\text{e-}3\}$ for $\ell_{\infty}$ attacks. We observe that PGD~\cite{pgd2018} is efficient in computing an imperceptible perturbations for different ranges of step-size $\alpha$. 
	}}
	\label{tbl:main}
	\label{fig:sup_indirect_city}
\end{figure*}


\providecommand{\localwidth}{}
\renewcommand{\localwidth}{\linewidth}

\providecommand{\localheight}{}
\renewcommand{\localheight}{2cm}

\begin{figure*}[hbt!]
	\centering
	\includegraphics[width=\localwidth]{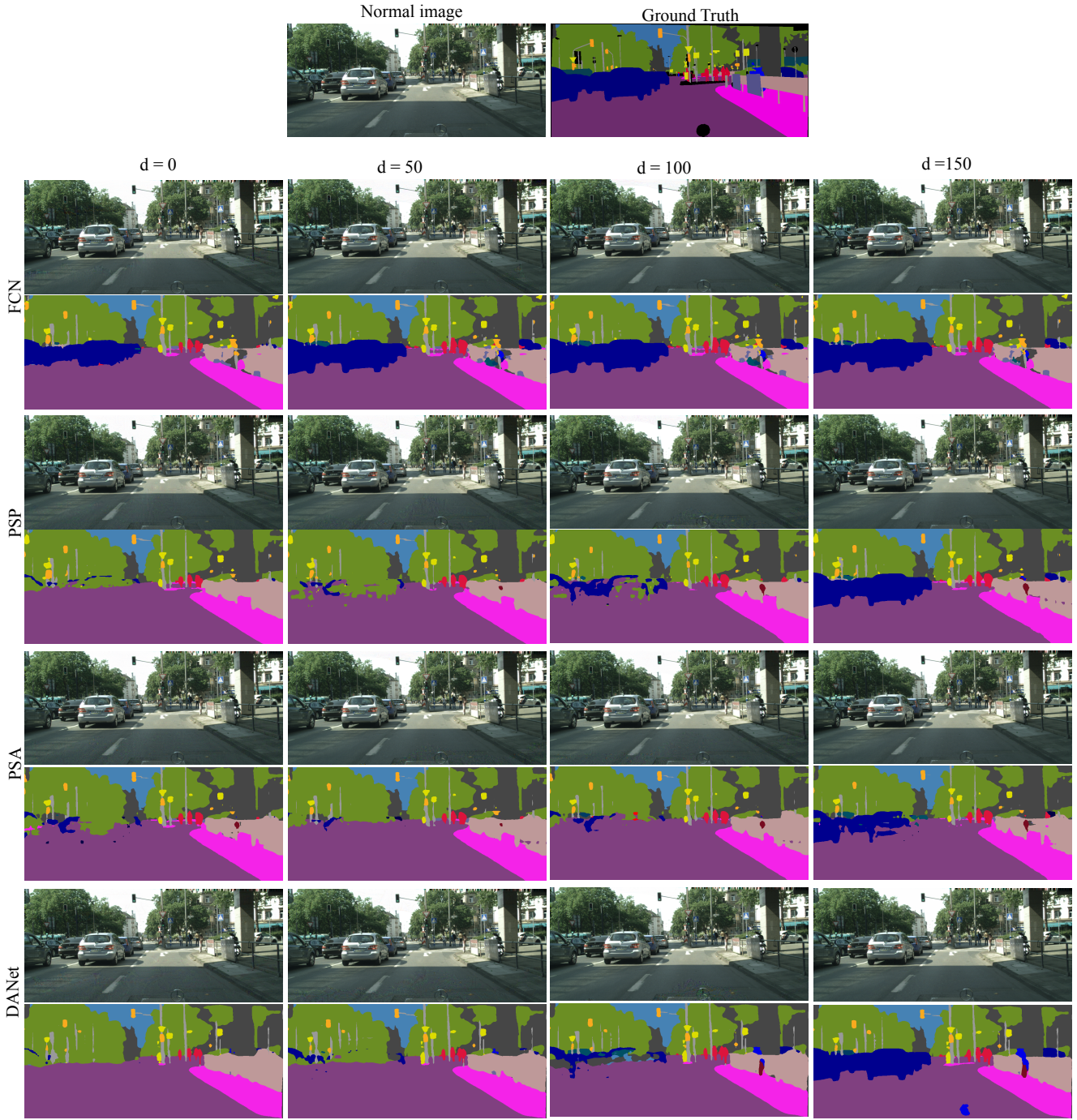}
		\vspace*{0.2cm}
	\caption{{\bf Indirect Local attack} on different networks with perturbations at least $d$ pixels away from any dynamic class. In most cases, FCN~\cite{fcn2015} is not effected to indirect attacks while PSANet~\cite{psanet2018}, PSPNet~\cite{psp2017} and DANet~\cite{danet2019} are effected due to larger contextual dependencies for prediction. } 
	\label{fig:sup_boundary_city}
\end{figure*}


\providecommand{\localwidth}{}
\renewcommand{\localwidth}{\linewidth}

\providecommand{\localheight}{}
\renewcommand{\localheight}{2cm}

\begin{figure*}[hbt!]
	\centering
	\includegraphics[width=\localwidth]{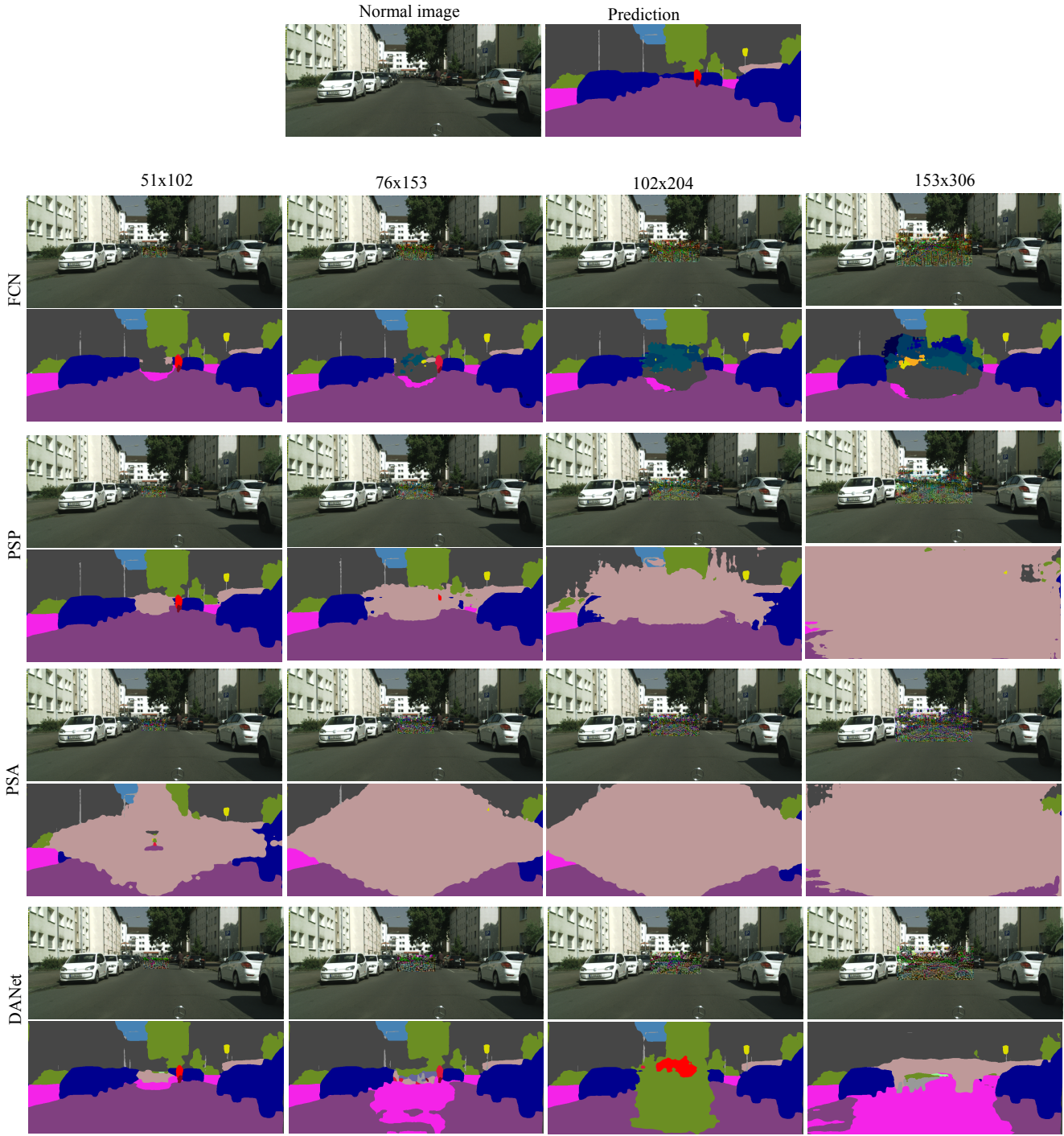}
		\vspace*{0.2cm}
	\caption{{\bf  Universal Local Attacks} on segmentation networks. The degradation in FCN~\cite{fcn2015} is limited to the attacked area, whereas for context-aware networks, such as PSPNet~\cite{psp2017}, PSANet~\cite{psanet2018}, DANet~\cite{danet2019}, it extends to far away regions.}
	\label{fig:sup_univ_city}
\end{figure*}


\providecommand{\localwidth}{}
\renewcommand{\localwidth}{\linewidth}

\providecommand{\localheight}{}
\renewcommand{\localheight}{2cm}

\begin{figure*}[hbt!]
	\centering
	\includegraphics[width=\localwidth]{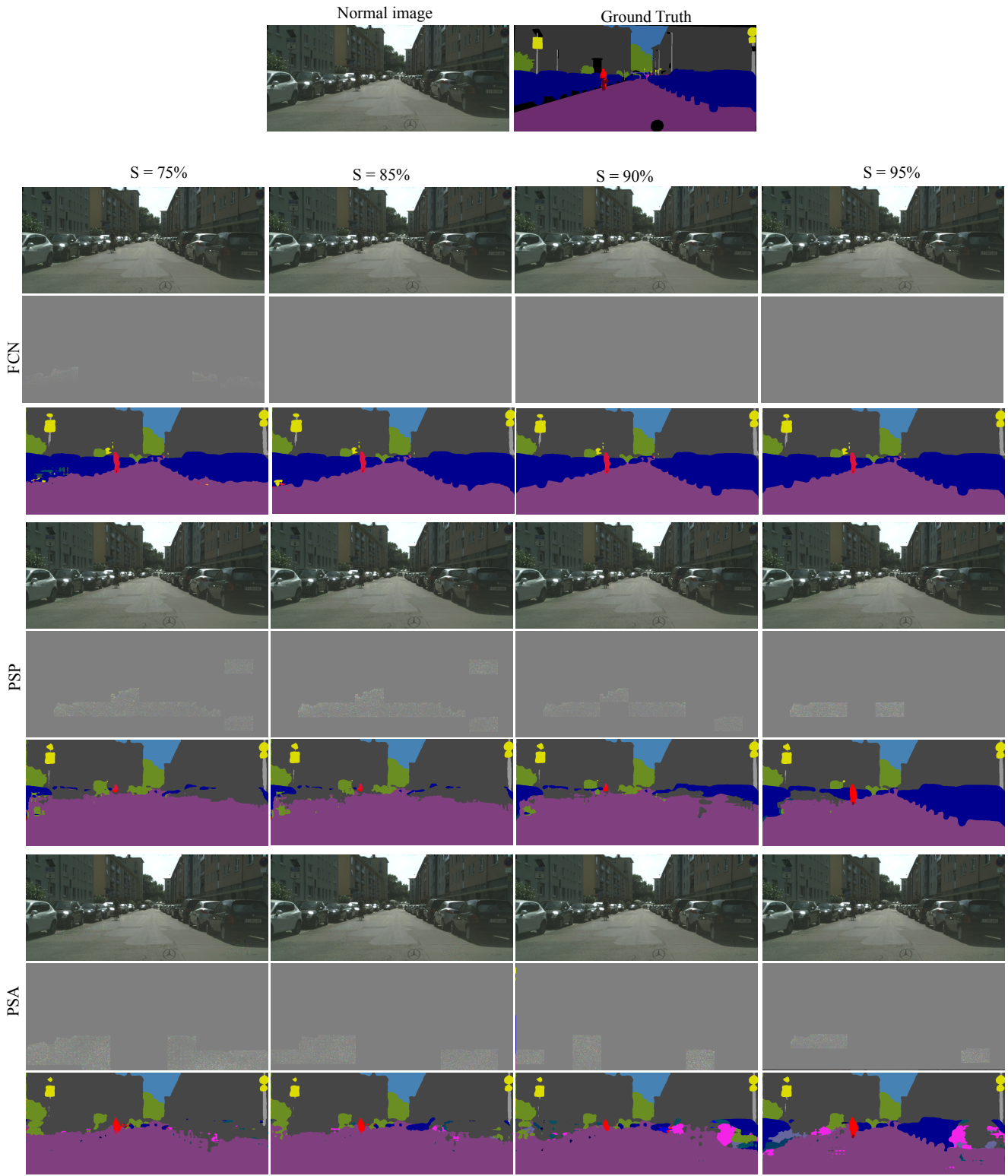}
		\vspace*{0.2cm}
	\caption{{\bf Adaptive Indirect Local Attacks on Cityscapes with different networks by tuning the sparsity levels}. We observe that PSPNet~\cite{psp2017} and PSANet~\cite{psanet2018} are most vulnerbale to adaptive indirect local attacks even with perturbations with high levels of sparsity while FCN~\cite{fcn2015} is least effected.  }
	\label{fig:sup_adaptive_city}
\end{figure*}


\providecommand{\localwidth}{}
\renewcommand{\localwidth}{\linewidth}

\providecommand{\localheight}{}
\renewcommand{\localheight}{2cm}

\begin{figure*}[hbt!]
	\centering
	\includegraphics[width=0.8\localwidth]{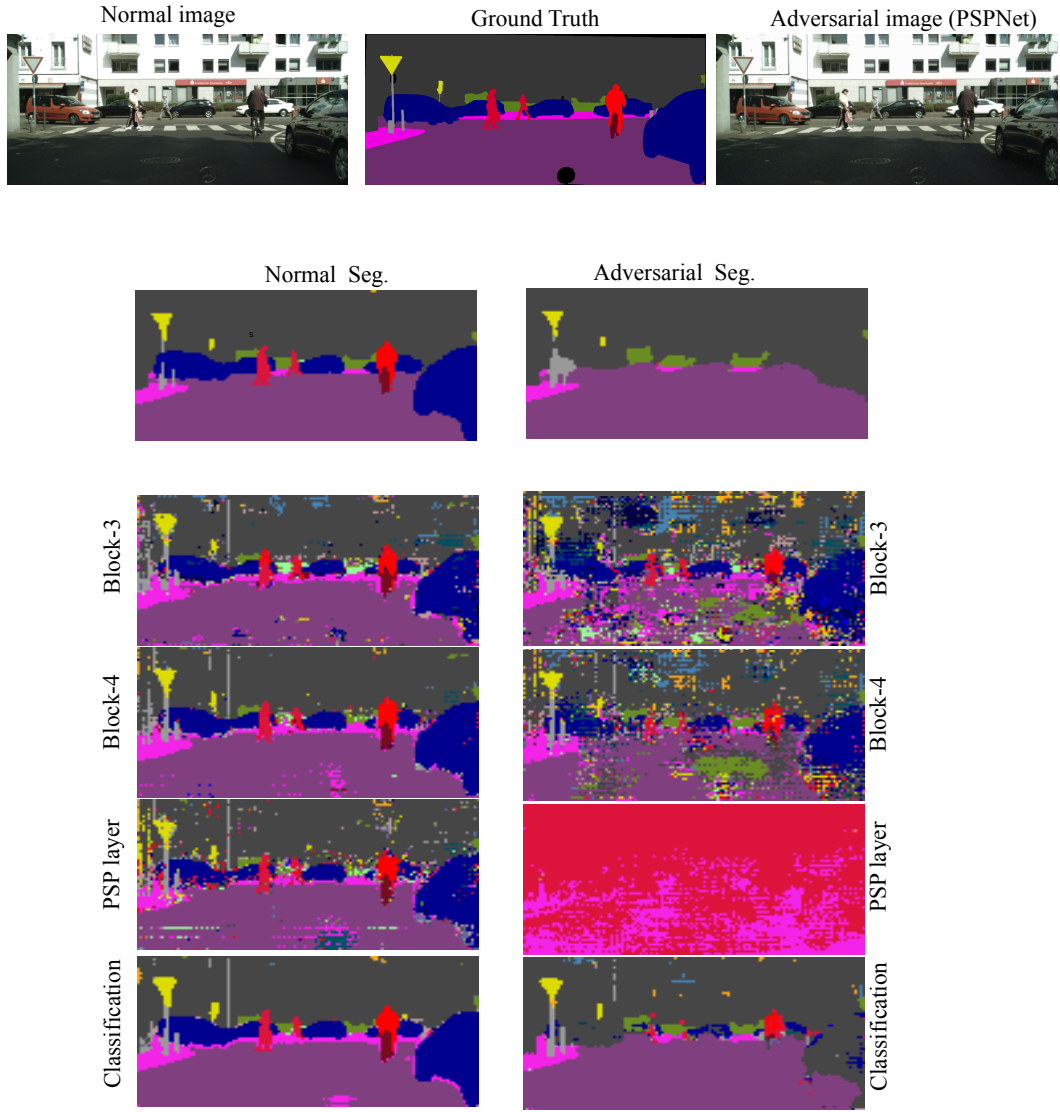}
		\vspace*{0.2cm}
	\caption{{\bf Visualizing internal subspaces of normal and adversarial samples of Cityscapes with PSPNet~\cite{psp2017}}. For each spatial location of the extracted feature map at layer $\ell$, we assign the label of nearest pre-trained class-conditional distribution computed using mahalanobis distance. As shown in figure, the nearest cluster label almost looks same as the predicted label map for clean samples however for adversarial samples, the nearest cluster moves towards the predicted adversarial label in final layers. Also,  in the context PSP layer, the nearest conditional distribution values are completely erroneous and far away from normal cluster assignments for adversarial samples.  
	}
	\label{fig:viz_psp_city}
\end{figure*}

\begin{figure*}[hbt!]
	\centering
	\includegraphics[width=0.8\localwidth]{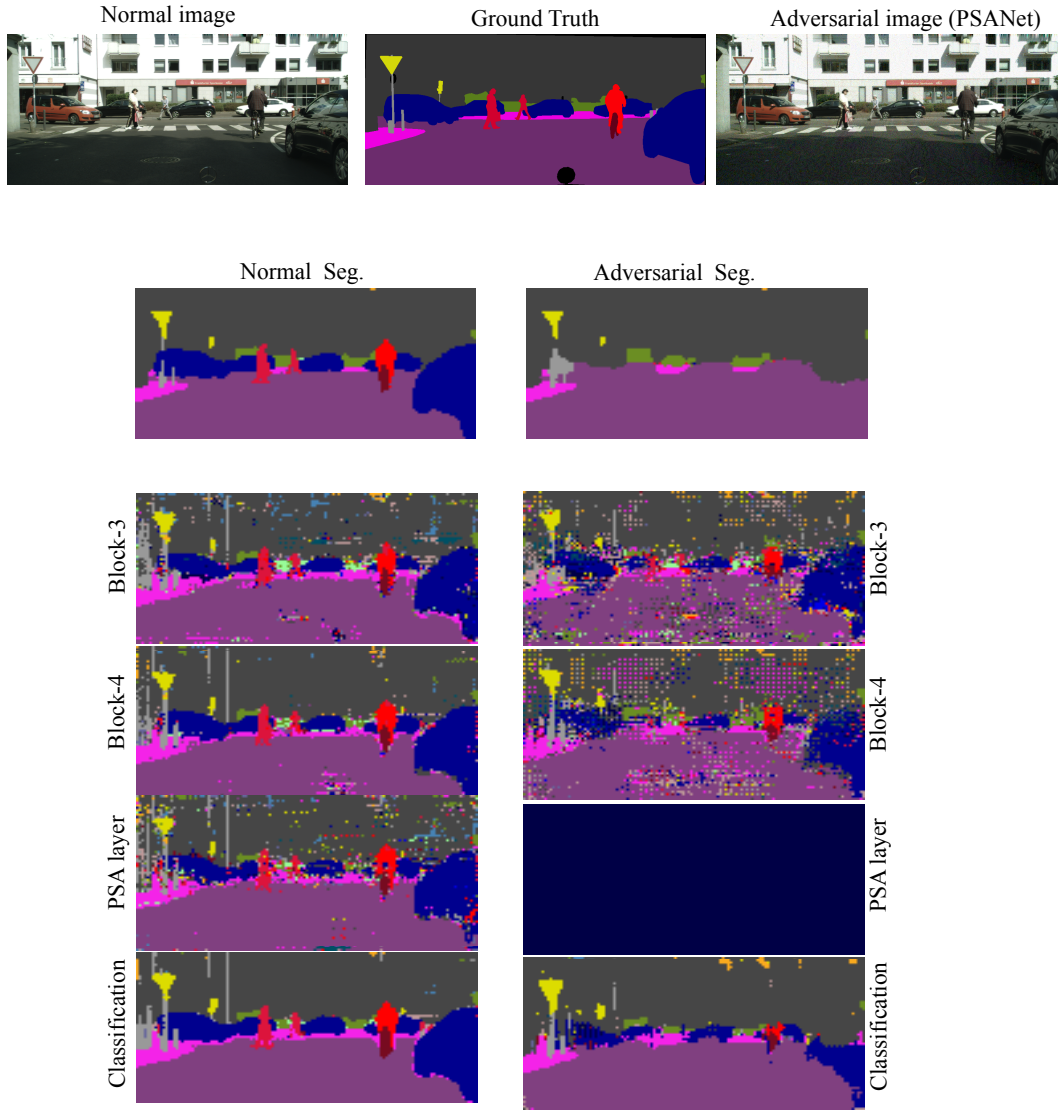}
		\vspace*{0.2cm}
	\caption{{\bf Visualizing internal subspaces of normal and adversarial samples of Cityscapes with PSANet~\cite{psanet2018}}. For each spatial location of the extracted feature map at layer $\ell$, we assign the label of nearest pre-trained class-conditional distribution computed using mahalanobis distance. As shown in figure, the nearest cluster label almost looks same as the predicted label map for clean samples however for adversarial samples, the nearest cluster moves towards the predicted adversarial label in final layers. Also,  in the context PSA layer, the nearest conditional distribution values are completely erroneous and far away from normal cluster assignments for adversarial samples.  
	}
	\label{fig:viz_psa_city}
\end{figure*}


\providecommand{\localwidth}{}
\renewcommand{\localwidth}{\linewidth}

\providecommand{\localheight}{}
\renewcommand{\localheight}{2cm}

\begin{figure*}[hbt!]
	\centering
	\includegraphics[width=0.6\localwidth]{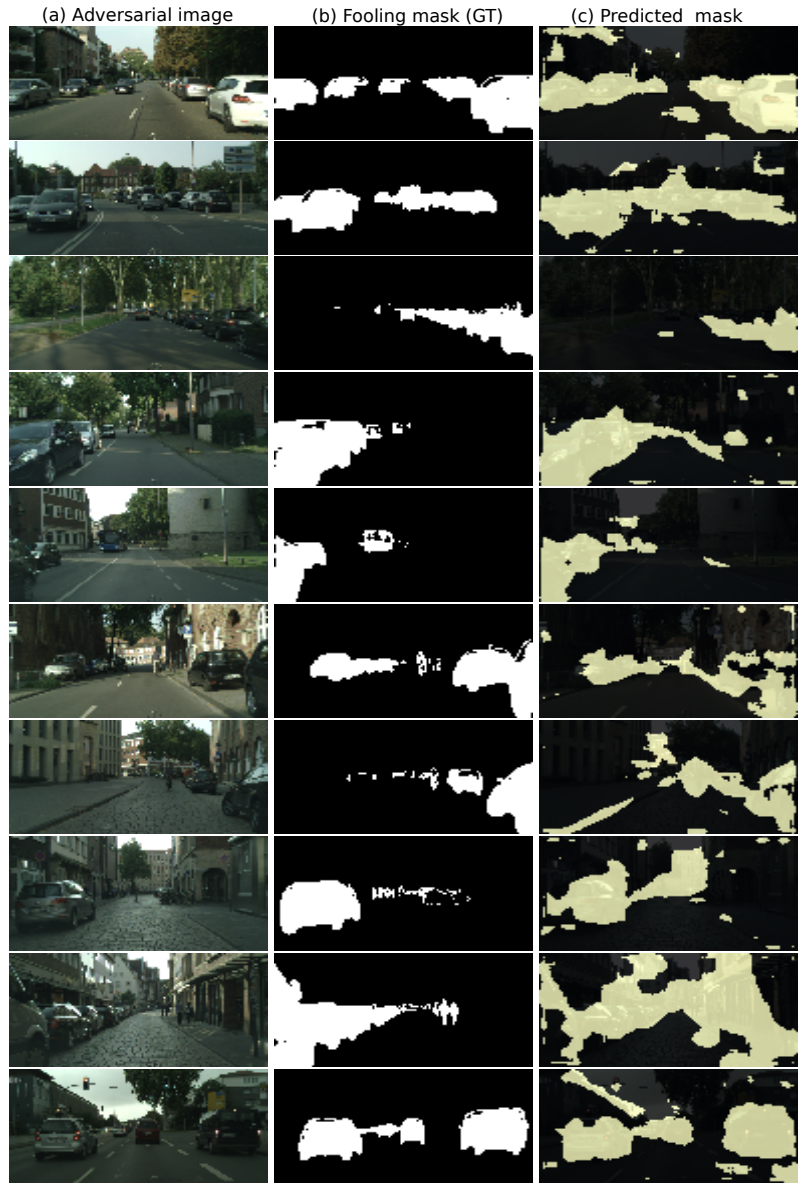}
		\vspace*{0.3cm}
	\caption{{\bf Visualization of attack detection} at pixel level by adaptive Indirect Local Attacks on Cityscapes with PSANet~\cite{psanet2018}. The first column shows the adversarial image, second column shows the ground truth fooling positions and the third column the predicted fooling positions.}\label{fig:detect_city1}
\end{figure*}

\subsection{PASCAL VOC Experiments}

\kn{Table~\ref{tbl:apative_full_voc} shows the robustness of FCN~\cite{fcn2015} local indirect attacks than PSANet~\cite{psanet2018}. For example, at Sparsity level of $95\%$, FCN~\cite{fcn2015} has success rate of $13\%$ as compared to  $68\%$ for PSANet.}

\begin{table*}[t]
	\begin{center}
		{ 
			\resizebox{0.6\linewidth}{!}{%
					\begin{tabular}{@{}cc@{\hskip 0.3in}cc@{\hskip 0.3in}ccc@{\hskip 0.3in}cc@{}}
					\midrule
					\multirow{2}{*}{Networks}   
					& 	\multirow{2}{*}{ Sparsity }    
					&  \multicolumn{2}{c}{\centering mIoU}  
					& \multicolumn{3}{c}{\centering mASR} 
					& \multicolumn{2}{c}{Norm of $\delta$}  
					\\  \cmidrule(l{-5pt}r{16pt}){3-4} \cmidrule(l{-5pt}r{15pt}){5-7}  \cmidrule(l{-2pt}r){8-9}
					& 
					& $\text{mIoU}_{\mathbf{u}}^{\bfs}$  
					& $\text{mIoU}_{\mathbf{t}}^{\bfs}$  
					& $\text{mASR}_{\mathbf{u}}^{\bp}$ 
					& $\text{mASR}_{\mathbf{u}}^{\bfs}$ 
					& $\text{mASR}_{\mathbf{t}}^{\bfs}$ 
					& $\ell_{\infty}$-norm 
					& $\ell_{2}$-norm \\
					\midrule
 					
 					\multirow{4}{*}{\begin{tabular}[c]{@{}c@{}} FCN~\cite{fcn2015} \end{tabular}} 
 				& $75\%$ &0.50&0.32&100\%&35\%&32\%&0.14 & 2.40\\
 				& $85\%$ &0.58&0.27&100\%&30\%&27\%&0.13 &2.15\\
 				& $90\%$ &0.66&0.22&100\%&24\%&22\%&0.12& 1.91\\
 				& $95\%$ &0.80&0.12&100\%&13\%&13\%& 0.11 & 1.37\\
 				\midrule
					\multirow{4}{*}{\begin{tabular}[c]{@{}c@{}} PSANet~\cite{psanet2018} \end{tabular}} 
				& $75\%$ &0.29&0.68&99\%&70\%&68\%&0.07&1.77 \\
			& $85\%$ &0.22&0.78&98\%&79\%&78\%&0.07& 1.93 \\
			& $90\%$ &0.20&0.80&98\%&82\%&80\%&0.08 & 2.21\\
			& $95\%$ &0.30&0.69&99\%&70\%&68\%&0.13 &2.81\\
 	 					\bottomrule
				\end{tabular}
			}
		}
\end{center}
	\caption{ {\small \textbf{Adaptive Indirect Local Attacks} on PASCAL VOC. We observe that PSANet~\cite{psanet2018} is more vulnerable to local adaptive attacks than FCN~\cite{fcn2015}.}}
	\label{tbl:apative_full_voc}
	\vspace*{-0.6cm}
\end{table*}


\providecommand{\localwidth}{}
\renewcommand{\localwidth}{\linewidth}

\providecommand{\localheight}{}
\renewcommand{\localheight}{2cm}

\begin{figure*}[hbt!]
	\centering
	\includegraphics[width=0.6\localwidth]{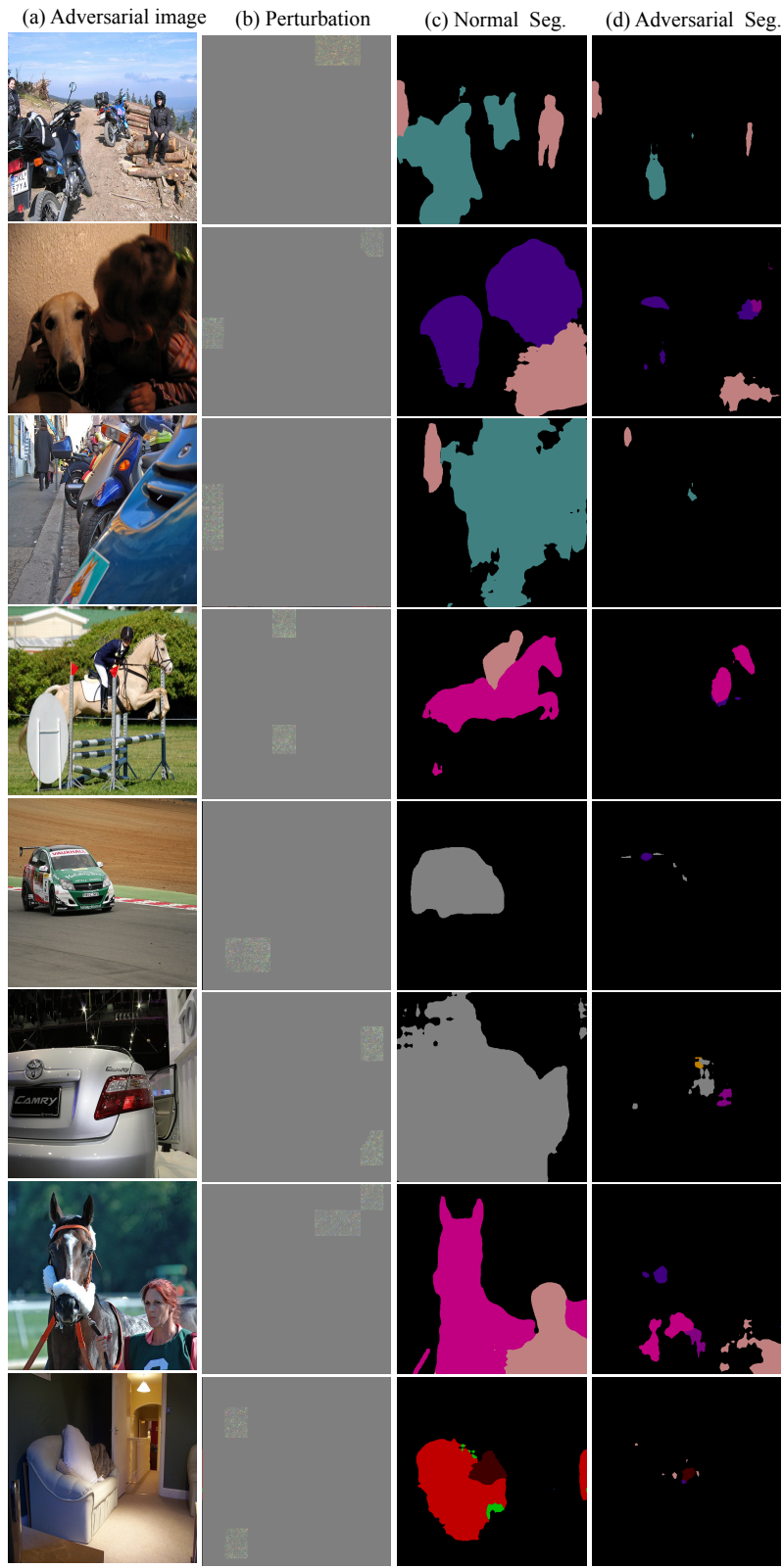}
	\vspace*{0.2cm}
	\caption{{\bf Adaptive Indirect Local Attacks} on PASCAL VOC  with PSANet~\cite{psanet2018}. The first column shows an adversarial image {\bf (a)} perturbed an imperceptible noise {\bf (b)} at local background regions  mis-classifies the foreground label  in normal segmentation map {\bf (c)} as the  background ones shown in {\bf (d)}. }
	\label{fig:sup_adaptive_voc}
\end{figure*}

\subsubsection{Qualitative Results on PASCAL VOC}

Figure~\ref{fig:sup_adaptive_voc}  shows the results of adaptive local attacks on PSANet~\cite{psanet2018} at Sparsity $95\%$.


\end{document}